\documentclass[journal]{IEEEtran}
\usepackage{amsmath,amsfonts}
\usepackage{algorithmic}
\usepackage{algorithm}
\usepackage{array}
\usepackage[caption=false,font=normalsize,labelfont=sf,textfont=sf]{subfig}
\usepackage{textcomp}
\usepackage{stfloats}
\usepackage{url}
\usepackage{verbatim}
\usepackage{graphicx}
\usepackage{cite}
\usepackage{makecell}
\usepackage{threeparttable}
\usepackage{xcolor}
\usepackage{ulem}
\usepackage[absolute,overlay]{textpos}
\begin{document}

\title{Multi-Mode Pneumatic Artificial Muscles Driven by \\ Hybrid Positive-Negative Pressure}


\author{Siyuan Feng, Ruoyu Feng, and Shuguang Li\textsuperscript{*}
\thanks{Manuscript received February 24, 2025; revised August 19, 2025 and November 26, 2025; accepted February 7, 2026. This work was supported by the National Natural Science Foundation of China under Grant 52575025.}
\thanks{Siyuan Feng, Ruoyu Feng, and Shuguang Li are with the Department of Mechanical Engineering at Tsinghua University, Beijing, 100084, China. Shuguang Li is also with Beijing Key Laboratory of Transformative High-end Manufacturing Equipment and Technology, Tsinghua University, Beijing 100084, China.}
\thanks{\textsuperscript{*}Corresponding author: Shuguang Li (e-mail: lisglab@tsinghua.edu.cn).}
}




\maketitle

\begin{textblock*}{\textwidth}(-0.75cm,1cm)
\centering
\footnotesize
This work has been published in IEEE Transactions on Robotics. Link: https://doi.org/10.1109/TRO.2026.3666154
\end{textblock*}

\begin{textblock*}{\textwidth}(-3.9cm,1.3cm)
\centering
\footnotesize
Digital Object Identifier (DOI): 10.1109/TRO.2026.3666154
\end{textblock*}

\begin{textblock*}{\textwidth}(1.7cm,26.7cm)
\centering
\footnotesize
© 2026 IEEE.  Personal use of this material is permitted.  Permission from IEEE must be obtained for all other uses, in any current or future media, including reprinting/republishing this material for advertising or promotional purposes, creating new collective works, for resale or redistribution to servers or lists, or reuse of any copyrighted component of this work in other works.
\end{textblock*}



\begin{abstract}
Artificial muscles embody human aspirations for engineering lifelike robotic movements. This paper introduces an architecture for Inflatable Fluid-Driven Origami-Inspired Artificial Muscles (IN-FOAMs). A typical IN-FOAM consists of an inflatable skeleton enclosed within an outer skin, which can be driven using a combination of positive and negative pressures (e.g., compressed air and vacuum). IN-FOAMs are manufactured using low-cost heat-sealable sheet materials through heat-pressing and heat-sealing processes. Thus, they can be ultra-thin when not actuated, making them flexible, lightweight, and portable. The skeleton patterns are programmable, enabling a variety of motions, including contracting, bending, twisting, and rotating, based on specific skeleton designs. We conducted comprehensive experimental, theoretical, and numerical studies to investigate IN-FOAM's basic mechanical behavior and properties. The results show that IN-FOAM's output force and contraction can be tuned through multiple operation modes with the applied hybrid positive-negative pressure. Additionally, we propose multilayer skeleton structures to enhance the contraction ratio further, and we demonstrate a multi-channel skeleton approach that allows the integration of multiple motion modes into a single IN-FOAM. These findings indicate that IN-FOAMs hold great potential for future applications in flexible wearable devices and compact soft robotic systems.
\end{abstract}


\section{Introduction}
\IEEEPARstart{N}{atural} muscles possess remarkable mechanical properties, enabling animals to perform both dexterous and load-bearing tasks \cite{kier2007arrangement,deiringer2023functional}. To achieve competent capabilities like muscles, researchers have long sought to develop artificial muscles that can generate a wide range of motions, such as contraction, bending, and twisting \cite{mirvakili2018artificial}. Recent advances in artificial muscle techniques have explored a variety of actuation mechanisms \cite{zhang2019robotic}, including electric \cite{pelrine2000high,gu2017modeling,zhao2018compact}, thermal \cite{mirvakili2018artificial, madden2004artificial}, magnetic \cite{hu2018small,kim2019ferromagnetic}, and fluidic \cite{mosadegh2014pneumatic, higueras2021cavatappi,cui2021enhancing,chen2023morphological} actuation. Among them, fluid-driven artificial muscles are attractive for multiple practical applications due to their ease of manufacture, low cost, and versatile morphing abilities.

Fluid-driven artificial muscles primarily operate via positive or negative pressure. Positive pressure-driven artificial muscles can be fabricated from film materials, enabling a low-profile (whose thickness can be much smaller than the length or width when not actuated), lightweight, and compact design. They can produce large output forces as long as the driving pressure does not exceed their strength limitations. Established types like McKibben muscles \cite{chou1996measurement}, Pleated Pneumatic Artificial Muscles \cite{daerden2001concept}, and Pouch Motors \cite{niiyama2015pouch}, are generally limited in contraction ratio and motion programmability. Innovations like Paired Pouch Motors\cite{oh2019design}, Gusseted Pouch Motors\cite{jang2023design}, and series PAMs\cite{greer2017series} have increased the contraction ratio but still lack diverse motion programmability. More recently, X-PAMs \cite{feng2023x} achieve a high contraction ratio of up to 92.9\:\%, but rely on an X-crossing mechanism which sacrifices the low-profile advantage of thin-film materials. In contrast, vacuum-driven artificial muscles offer inherent safety (no bursting risk, minimal spatial expansion) and can achieve high contraction ratios ($>$90\:\%). These include VAMPs \cite{yang2016buckling}, FOAMs \cite{li2017fluid}, and origami-like silicone rubber artificial muscles \cite{jiao2019advanced}, and they exploit buckling or skin tension under vacuum. However, generally they require bulky 3D structures at the initial state to enable the contraction, which is different from the compact and low-profile designs of those driven by positive pressure.

\begin{figure*}
    \centering
    \includegraphics[width=1\textwidth]{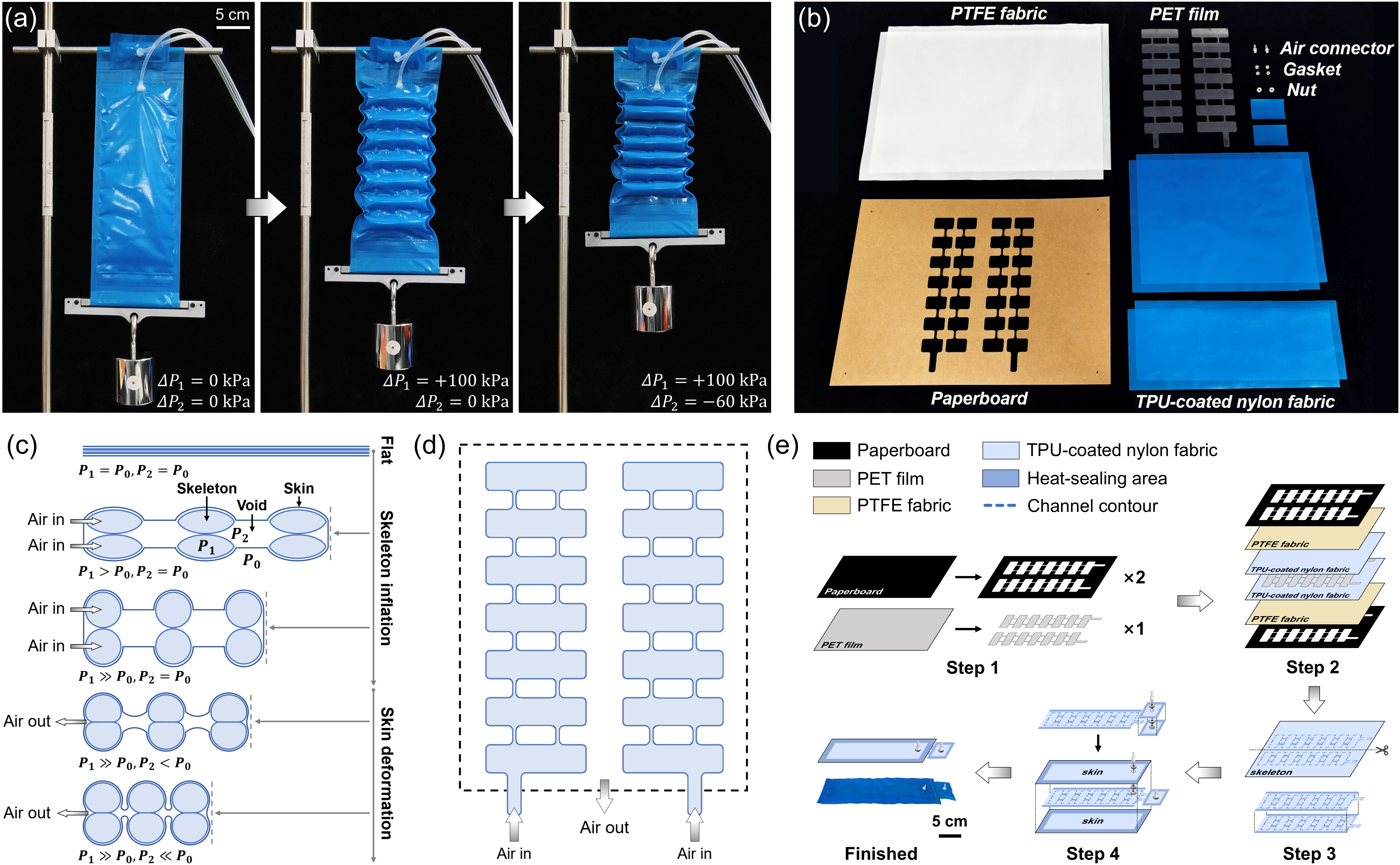}
    \caption{Design and fabrication of IN-FOAMs. (a) A linear IN-FOAM made of TPU-coated nylon fabrics. (b) Materials used to fabricate the actuator. The sheet materials have been patterned through engraving. (c) Working principle. The actuator is flat when not actuated. Skeleton inflation induces tension in the sheets to generate contraction, and voids appear between skeleton columns. After applying a vacuum to the voids, the difference in pressure relative to the atmosphere induces tension in the skin and drives the actuator to contract further. (d) A pneumatic circuit design of IN-FOAMs. (e) Fabrication process. IN-FOAMs can be fabricated rapidly in four steps: (step 1) pattern the sheets, (step 2) stack and heat-press the sheets, (step 3) cut and heat-seal layers of the skeleton, (step 4) add pneumatic connectors and heat-seal the skin.}

    \label{figPrinciple}
\end{figure*}

Some single-chamber soft actuators can be driven by either positive pressure or negative pressure. For example, bellows-enclosed\cite{drotman2018application}, origami-enclosed\cite{oh2024hybrid
}, and pleated\cite{de20223d} soft actuators are capable of extension and contraction under positive and negative pressure, respectively. Soft actuators employing simultaneous positive and negative pressures in dual chambers have also been developed, demonstrating enhanced functionality, including antagonistic force control\cite{usevitch2018apam}, increased bending force\cite{fatahillah2020novel}, large braking force\cite{jang2022positive}, and amplified pressure difference\cite{coutinho2023hyperbaric}. By regulating the pressures in inner and outer chambers, an antagonistic pneumatic artificial muscle can generate pushing or pulling forces and independent stiffness control\cite{usevitch2018apam}. Driven by combined positive and negative pressures, a soft bending actuator provides a larger bending force at lower bending angles than that driven by a single pressure source\cite{fatahillah2020novel}. A soft linear brake can produce large braking forces and adjust its behavior through the combination of positive and negative pressures \cite{jang2022positive}. The limited contraction ratio and bulky structures of these actuators can still be improved.

Integrating the benefits of positive pressure-driven and vacuum-driven artificial muscles into a single actuator that simultaneously achieves a low-profile lightweight structure, a high contraction ratio, and versatile motion capabilities remains challenging. This issue is particularly critical for space-constrained applications (e.g., wearables, pipe or cave exploration), which require compact designs alongside advanced actuation performance. Despite significant advances in fluid-driven artificial muscles, existing low-profile actuators often exhibit limited contraction ratios or motion modes. Conversely, actuators with high contraction ratios or multi-mode motions tend to be bulky, indicating a need for designs that combine all three desirable features.


To bridge this gap, we introduce inflatable fluid-driven origami-inspired artificial muscles (IN-FOAMs), aiming to combine the advantages of both positive pressure-driven and negative pressure-driven artificial muscles. Driven by hybrid positive-negative air pressure, IN-FOAMs utilize inflatable skeletons to achieve ultrathin initial configurations and use the negative air pressure to generate further deformations. This hybrid approach integrates the low-profile form factor of thin-film pneumatic artificial muscles and the high contraction ratio of vacuum-driven artificial muscles.
The contributions of our work are as follows:
\begin{itemize}
    \item[\textbullet] \textbf{Ultra-thin, flexible and portable design:} IN-FOAMs are fabricated from heat-sealable flexible sheets, resulting in inherent flexibility, light-weight, and an ultrathin, low-profile resting state — a significant advantage over conventional FOAMs. This allows IN-FOAMs to be rolled or folded for portability.
    \item[\textbullet] \textbf{High contraction ratio in a pouch form:} In addition to contraction by inflation (like pouches), IN-FOAMs integrate negative pressure actuation and multilayer skeleton design, providing a kind of pouch-shaped pneumatic artificial muscle with a maximum contraction ratio of more than 50\:\%.
    \item[\textbullet] \textbf{Enhanced motion programmability:} Programmable patterns of the inflatable skeleton channels enable diverse motions, including contraction, bending, twisting, and rotation, enhancing the motion programmability of pouch-shaped pneumatic artificial muscles. We further demonstrate a multi-mode IN-FOAM integrating several motion modes in a single compact actuator.
    \item[\textbullet] \textbf{Modeling and simulation method:} We develop a theoretical model for linear IN-FOAMs and utilize finite element simulation to analyze the working principle of the combined positive and negative pressure actuation. We demonstrate the feasibility of numerical prediction of IN-FOAMs' motions, thus establishing a foundation for computational design.
\end{itemize}

\begin{figure*}
    \centering
    \includegraphics[width=1\textwidth]{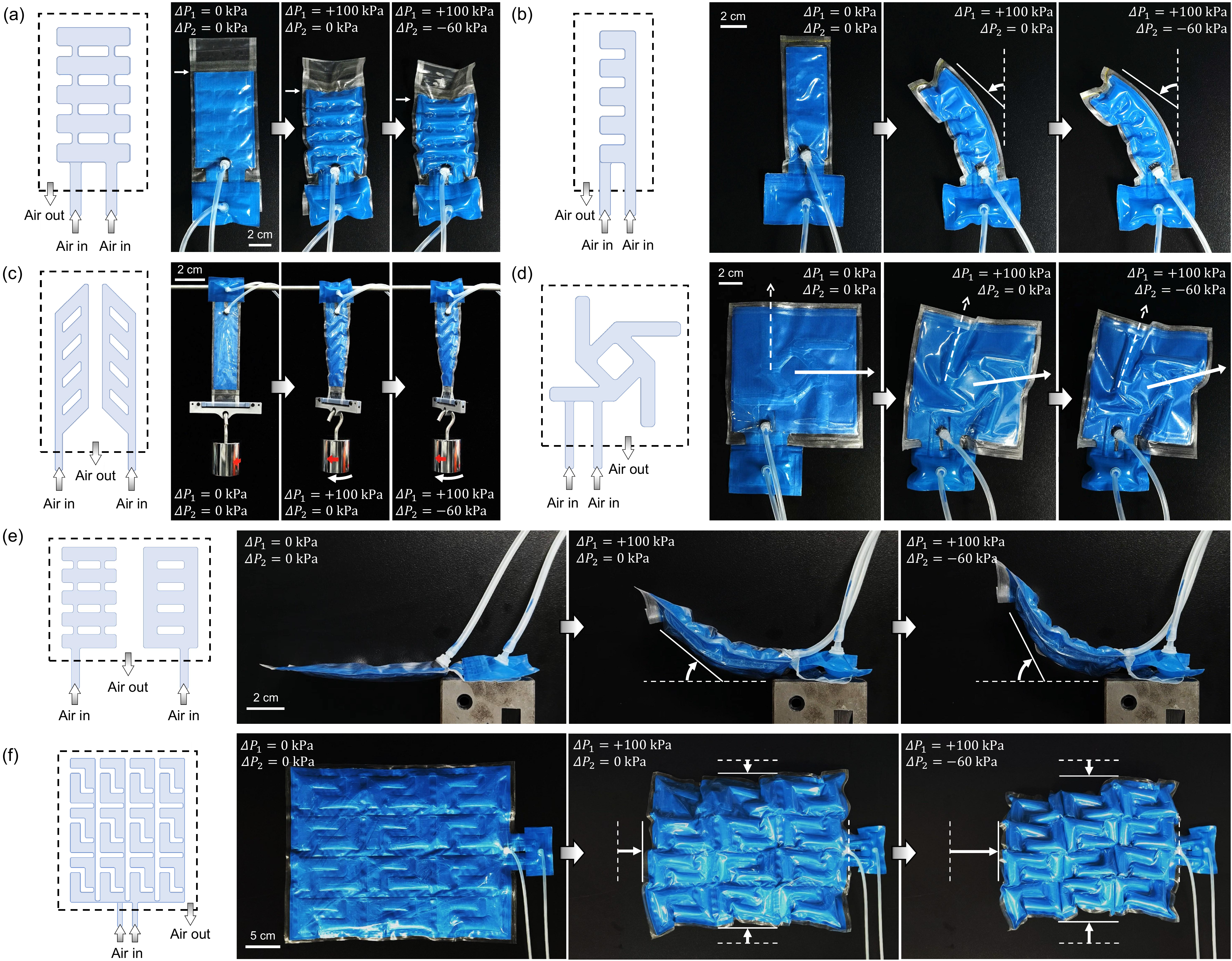}
    
    \caption{Various motions performed by IN-FOAMs. The actuators' skeletons are made of TPU-coated nylon fabrics (thickness: 0.09 mm), and the skins are made of transparent TPU film (thickness: 0.1 mm). Without actuation, the actuators are flat. (a) A linear IN-FOAM with a parallel-pouch skeleton (size: $165\times60\times0.65\: \rm{mm^3}$, in the initial state; weight: 6.26 g) contracts 17 mm driven by positive pressure and further contracts 7 mm under the negative-pressure actuation. (b) A parallel-pouch skeleton with an inflated beam at its right side enables the actuator (size: $145\times60\times0.65\: \rm{mm^3}$, in the initial state; weight: 4.31 g) to bend 47° leftwards driven by positive pressure and further bend 8° under the negative-pressure actuation. (c) With inflated beams to constrain contraction of the bottom skeleton, the actuator (size: $145\times60\times0.65\: \rm{mm^3}$, in the initial state; weight: 5.97 g) with a parallel-pouch skeleton is allowed to bend 39° upwards and further bend 23° under positive and negative pressure actuation, respectively. (d) By an anti-symmetric arrangement of oblique parallel pouches in the upper and bottom skeleton, the actuator (size: $215\times60\times0.65\: \rm{mm^3}$, in the initial state; weight: 6.65 g) twists 79° and further twists 13° under the positive and negative pressure actuation, respectively. (e) A flasher-origami-like skeleton makes the actuator (size: $160\times112\times0.65\: \rm{mm^3}$, in the initial state; weight: 10.34 g) rotate nearly 23° driven by positive pressure and further rotate 10° under the negative-pressure actuation. (f) By patterning the skeleton with L-shaped units, the actuator (size: $380\times260\times0.65\: \rm{mm^3}$, in the initial state; weight: 58.80 g) generates a length decrease of 36 mm and a width decrease of 41 mm under the positive pressure actuation; under the negative pressure actuation, the length and the width further decrease by 34 mm and 13 mm, respectively.}

    \label{figMotions}
\end{figure*}

\section{Design and Fabrication}
\subsection{Working Principle}
Different from the structure of FOAMs\cite{li2017fluid}, which consists of a skin and a skeleton, the proposed IN-FOAM has an inflatable skeleton, as shown in Fig. \ref{figPrinciple}(a). Here, we take the linear IN-FOAM as an example (one pneumatic circuit design is shown in Fig. \ref{figPrinciple}(c)), which performs linear contraction and has a two-layer skeleton. The skeleton can have multiple patterned chambers; for the linear IN-FOAM, the pattern of pouches is used, like pouch motors\cite{niiyama2015pouch}. Without actuation, the artificial muscle is flat, and its thickness is approximately the total thickness of the laminates. In the operation process, the skeleton is inflated with a positive pressure {$\Delta P_1$}, generating tension and contraction. Upon inflation of the skeleton, voids appear between a series of pouches. Then, a vacuum is applied to these voids. The pressure difference between the external atmosphere and the voids, {$\Delta P_2$}, induces tension in the skin and generates a further contraction.

\subsection{Fabrication}
The IN-FOAMs can be manufactured through heat-pressing and heat-sealing processes using low-cost heat-sealable airtight materials (e.g., nylon fabric coated with two-sided thermoplastic polyurethane (TPU) films). As shown in Fig. \ref{figPrinciple}(d), the fabrication process of the artificial muscles can be divided into four steps. Step 1 is to pattern the sheets through engraving. Step 2 is to heat-press different materials. Polyethylene terephthalate (PET) films are used as masks to prevent adhesion of the TPU-coated nylon fabric, and in this way, the fluidic channels can be created. Paperboards are used as a stamp to transmit the heat and load intensively into the bonding areas during the heat-pressing process. Polytetrafluoroethylene (PTFE) fabric sheets are used to prevent the adhesion of the paperboards and the TPU-coated nylon fabric, which is used as the skeleton material. Step 3 is to cut the skeleton into two layers and bond their edges. Step 4 is to add air inlet and outlet connectors and heat-seal the skin wrapping the skeleton. Besides TPU-coated nylon fabrics, other heat-sealable airtight sheet materials can be used to fabricate IN-FOAMs through the aforementioned method.

In this work, primary materials include N20D nylon fabric with two-sided TPU coating (Jiaxing Yingcheng Textile Co., Ltd., China), PET film (Shanghai Youshu Trading Co., Ltd., China), air connector (Guangzhou Sanglian Electronic Technology Co., Ltd., China), nylon nut (Shenzhen Bairuite Fastener Technology Co., Ltd., China), and silicone tubes (Nanjing Runze Fluid Control Equipment Co., Ltd., China). To fabricate IN-FOAMs, the heat-press is conducted for 50 seconds under a pressure of 350 kPa and a temperature of 200 ℃. After the heat-press, a small amount of plastic strain remains in IN-FOAMs, leading to slight length and thickness change compared to the raw materials.

\subsection{Prototypes with Programmable Motions}
The output motions of IN-FOAMs are programmed in the skeleton patterns. A skeleton pattern can be regarded as a combination of multiple line segments. The direction of tension induced by skeleton inflation is perpendicular to the longitudinal direction of these segments, and the direction of tension in the skin under the vacuum depends on the arrangement of the adjacent skeleton segments. Therefore, the tension distribution during the operation is determined by the skeleton segments and their configuration. A variety of motions can be produced by varying the skeleton design. Linear contraction is the most basic among these motions. It can be achieved by a skeleton in the form of parallel pouches (Fig. \ref{figMotions}(a)). Bending motion can be achieved by parallel pouches when one of its sides is constrained, and this is realized through an inflated beam (Fig. \ref{figMotions}(b)). The asymmetric contraction makes the actuator bend towards the unconstrained side. A skeleton with inflated beams in the bottom layer can constrain the contraction of the bottom layer and produce an upward bending (Fig. \ref{figMotions}(c)). Torsion can be produced by an anti-symmetric arrangement of the upper and bottom skeleton segments, generating torque in the longitudinal direction of the IN-FOAM during actuation. Then the skeleton segments are twisted under the torque (Fig. \ref{figMotions}(d)). A skeleton with a flasher-origami-like pattern can induce a torque and produce a rotation of 33° (Fig. \ref{figMotions}(e)). A skeleton with L-shaped units can achieve a 2D contraction (Fig. \ref{figMotions}(f)). Pneumatic circuit designs of the IN-FOAMs mentioned above are shown in Fig. \ref{figMotions}.

While programmable skeleton patterns allow IN-FOAMs to generate various motions, the applied negative pressure enables the actuators to magnify these motions on the basis of positive-pressure actuation, as shown in Fig. \ref{figMotions}. For the linear IN-FOAM, the negative pressure makes the maximum contraction ratio ($\sim$ 43\:\%) higher than the actuators only with inflated chambers, such as Pouch Motors ($\sim$ 28\:\%)\cite{niiyama2015pouch} and Peano Muscles ($\sim$ 17\:\%)\cite{veale2016characterizing}. This is thoroughly investigated in the next section.

\section{Modeling}
\begin{figure*}
    \centering
    \includegraphics[width=1\textwidth]{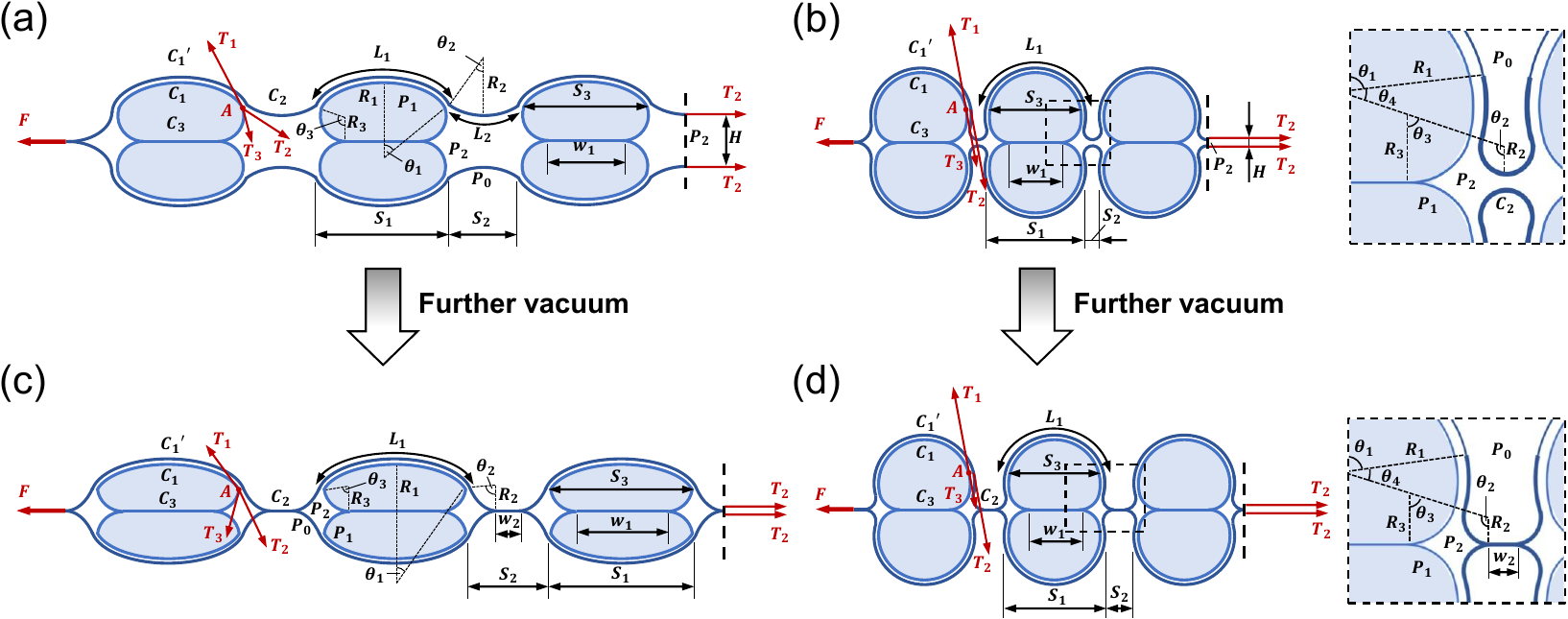}
    
    \caption{Illustration of the modeling for the linear IN-FOAM. Cross-sectional Views of the FE simulation results are shown in Fig. \ref{figSimulation}(a-d). (a) Model A: the skin between skeleton columns has no contact with the skeleton under the heat-pressed boundary (denoted by point A). (b) Model B: the skin contacts with the skeleton under point A as the linear IN-FOAM contracts further. (c) Model C: when the vacuum is high enough, the upper and bottom skin between skeleton columns come into contact on the basis of Model A. (d) Model D: when the vacuum is high enough, the upper and bottom skin between skeleton columns come into contact on the basis of Model B.}

    \label{figModel}
\end{figure*}

The modeling is based on the plane-strain theories of pressurized membranes \cite{thbaut2023pres}, considering the nonlinear deformation, geometric constraints, and contacts. It can be assumed that, the cross-section of the actuator is connected by the arcs with piecewise constant curvatures, and the straight lines at the contact regions (Fig. \ref{figModel}). The arcs $C_1$ and $C_3$ represent the two surfaces of each pouch. $C_1^{'}$ is part of the skin between the two ends of each pouch. We assume that, the skin and the skeleton between the adjacent pouch columns are bonded without any separation, and $C_2$ stands for this part of the sheet. $C_1$ and $C_1^{'}$ are also assumed to be bonded. Point {\it{A}} stands for the heat-pressed boundary of each pouch. Under the negative pressure, the upper and lower layers of the skeleton are pushed towards each other, forming a flat contact region with the length of $w_1$. When the negative pressure or the contraction is large enough, the skin between pouch columns comes into contact under pressure difference, forming different kinds of geometric relations. These agree with the FE simulations (Fig. \ref{figSimulation}(a-d)).

At the beginning of contraction, as shown in Fig. \ref{figModel}(a), the geometric equations can be established for a single pouch and its adjacent skin:
\begin{equation}\label{eqn-st1}
L_1=2R_1 \theta_1,
\end{equation}
\begin{equation}\label{eqn}
L_2=2R_2 \theta_2,
\end{equation}
\begin{equation}\label{eqn}
L_3=2R_3 \theta_3+w_1,
\end{equation}
\begin{equation}\label{eqn}
S_1=2R_1 \sin\theta_1,
\end{equation}
\begin{equation}\label{eqn}
S_2=2R_2 \sin\theta_2,
\end{equation}
\begin{equation}\label{eqn}
S_3=2R_3 \sin\theta_3+w_1,
\end{equation}
\begin{equation}\label{eqn}
S_1=S_3,
\end{equation}
where $L_1,L_2,L_3$ are the lengths of $C_1,C_2,C_3$, $R_1,R_2,R_3$ are the radii of the arcs $C_1,C_2,C_3$, $\theta_1,\theta_2,\theta_3$ are the central angles of these arcs, and $S_1,S_2,S_3$ are the spans of $C_1,C_2,C_3$.

The PET film used to fabricate IN-FOAMs is quite thin (including a PET layer of 8 $\mu m$ and an adhesive layer of 12 $\mu m$) and highly flexible, with a length ($\sim$ 20 mm for each skeleton pouch) that is about 1,000 times the thickness. The length-to-thickness ratios of the fabrics used are also high ($\sim$ 222 for the skeleton material and $\sim$ 118 for the skin material). So we consider the materials as membranes, and our analysis incorporates only tensile stiffness, disregarding bending stiffness. Based on Laplace's law, the tensions are calculated as
\begin{equation}\label{eqn-st5}
T_1={\Delta}P_1R_1 W=(P_1-P_0 ) R_1 W,
\end{equation}
\begin{equation}\label{eqn}
T_2=-{\Delta}P_2R_2 W=(P_0-P_2 ) R_2 W,
\end{equation}
\begin{equation}\label{eqn}
T_3=(P_1-P_2 ) R_3 W,
\end{equation}
where $T_1,T_2,T_3$ are the tensions in $C_1,C_2,C_3, P_0,P_1,P_2$ are the atmospheric pressure, the applied positive pressure, and the applied negative pressure. $W$ is the actuator's width, respectively. For simplicity, a linear elastic deformation is assumed, and the tensions are proportional to the elongations:
\begin{equation}\label{eqn}
T_i=K_i (L_i-L_{i0} ),i=1,2,3,
\end{equation}
where $K_i$ is the tensile stiffness of $C_i$, and $L_{i0}$ is the initial lengths of $C_i$. Considering that the fabrication processes may lead to plastic strain, as well as a redundant length between the bonded ends, the actual $L_{i0}$ is slightly longer than its designed value, $L_{i0,\:designed}$. A factor $\delta$ is introduced to estimate $L_{i0}$ by $L_{i0}=(1+\delta)L_{i0,\:designed}$. $K_i$ can be calculated by
\begin{equation}\label{eqn}
K_i=(E_i t_i W)/L_{i0}, i=1,2,3,
\end{equation}
where $E_i$ is the elastic modulus of the material, and $t_i$ is the thickness of the sheet. 

In the static or quasi-static states, the tensions are balanced at point {\it{A}}:
\begin{equation}\label{eqn}
T_1 \cos\theta_1+T_3 \cos\theta_3=T_2 \cos\theta_2,
\end{equation}
\begin{equation}\label{eqn-ed5}
T_1 \sin\theta_1=T_2 \sin\theta_2+T_3 \sin\theta_3.
\end{equation}

The actuator's output force {\it{F}} can be calculated by applying the balance condition to the part of the linear IN-FOAM shown in Fig. \ref{figModel}(a) (cut along the dashed line, which is the center line of $C_2$):
\begin{equation}\label{eqn-st4}
 F=2T_2+(P_0-P_2 )HW-F_r,
\end{equation}
where $(P_0-P_2)HW$ is the force generated by the pressure difference, $H$ is the height of the cross-section along the dashed line, and $F_r$ is the contraction resistance from the material in the heat-pressed areas and heat-sealed areas. $H$ can be derived from the cross-section's geometry:
\begin{equation}\label{eqn}
H=2R_3 (1-\cos\theta_3 )-2R_2 (1-\cos\theta_2 ).
\end{equation}

We assume that the resistance $F_r$ is proportional to the contraction:
\begin{equation}\label{eqn}
F_r=k_r (S_{sum}-S_{sum,0}),
\end{equation}
where $k_r$ is the resistance coefficient and can be obtained through experimental data (Appendix \ref{appResistance}, Fig. \ref{figResistance}(a-c)), $S_{sum}$ is the length of IN-FOAM during contraction, and $S_{sum,0}$ is the initial length.

Assuming that the linear IN-FOAM has a skeleton with {\it{n}} columns of pouches, then its length is
\begin{equation}\label{eqn}
S_{sum}=nS_1+(n-1) S_2.
\end{equation}
With the initial length
\begin{equation}\label{eqn}
S_{sum,0}=nL_{10}+(n-1) L_{20},
\end{equation}
the contraction ratio of the linear IN-FOAM can be calculated as
\begin{equation}\label{eqn-ed1}
CR=(S_{sum,0}-S_{sum})/(S_{sum,0})\times100\:\%.
\end{equation}

By combining the Equations (\ref{eqn-st1}-\ref{eqn-ed1}), we can solve the relation between the output force and the contraction ratio, given the applied positive and negative pressures. Also, the blocked force of the linear IN-FOAM can be obtained by making $CR=0\:\%$. We denote the model above as Model A. Near the end of the contraction, there is a certain part of $C_2$ contacting with $C_3$, as shown in Fig. \ref{figModel}(b). This case is described by Model B, and the geometry equations are updated to 
\begin{equation}\label{eqn-st6}
L_1=2\theta_1 R_1,
\end{equation}
\begin{equation}\label{eqn}
L_2=2R_1 \theta_4+2R_2 \theta_2,
\end{equation}
\begin{equation}\label{eqn}
L_3=2R_3 \theta_3+2R_1 \theta_4+w_1,
\end{equation}
\begin{equation}\label{eqn}
S_1=2R_1 \sin\theta_1,
\end{equation}
\begin{equation}\label{eqn}
S_2=2R_1 (\sin\theta_3-\sin\theta_1 )+2R_2 \sin\theta_2,
\end{equation}
\begin{equation}\label{eqn}
S_3=2R_3 \sin \theta_3+2R_1 [\sin(\theta_3+\theta_4)-\sin \theta_3]+w_1,
\end{equation}
\begin{equation}\label{eqn}
S_1=S_3,
\end{equation}
where $\theta_4$ is the central angle of the arc that is the overlap of $C_2$ and $C_3$. From the geometry shown in Fig. \ref{figModel}(b), we can obtain Equations (\ref{eqn-st2}) and (\ref{eqn-st3}): 
\begin{equation}\label{eqn-st2}
\theta_1+\theta_3+\theta_4=\pi,
\end{equation}
\begin{equation}\label{eqn-st3}
\theta_2=\theta_1+\theta_4.
\end{equation}

The calculation of output force and contraction ratio is the same as Equations (\ref{eqn-st4}) and (\ref{eqn-ed1}). By solving Equations (\ref{eqn-st5}-\ref{eqn-st4}) and (\ref{eqn-st6}-\ref{eqn-st3}), we can get the output force at a certain contraction ratio at the end of contraction.

When the applied negative pressure is too high, the upper and lower layers of the skin contact with each other between the skeleton columns (Fig. \ref{figModel}(c)), and the contacted length is denoted as $w_2$. When this occurs, the geometric equations become 

\begin{equation}\label{eqn}
L_2=2R_2 \theta_2+w_2,
\end{equation}
\begin{equation}\label{eqn}
L_3=2R_3 \theta_3+w_1,
\end{equation}
\begin{equation}\label{eqn}
S_2=2R_2 \sin\theta_2+w_2,
\end{equation}
\begin{equation}\label{eqn}
S_3=2R_3 \sin\theta_3+w_1.
\end{equation}
In this case, the vertical height of $C_2$ is equal to that of $C_3$:
\begin{equation}\label{eqn}
R_2 (1-\cos\theta_2 )=R_3 (1-\cos\theta_3 ).
\end{equation}

The contraction ratio can be calculated using Equation (\ref{eqn-ed1}), and the output force is calculated by
\begin{equation}\label{eqn-st8}
F=2T_2-F_r.
\end{equation}

We denote the model in this case as Model C. With high negative pressure, Model B changes to Model D near the end of the contraction, as shown in Fig. \ref{figModel}(d). Compared to Model B, changes in the geometric equations include
\begin{equation}\label{eqn}
L_2=2R_1 \theta_4+2R_2 \theta_2+w_2,
\end{equation}
\begin{equation}\label{eqn-st7}
R_2 (1-\cos\theta_2 )=R_3 (1-\cos\theta_3 ),
\end{equation}
where Equation (\ref{eqn-st7}) means the vertical height of $C_2$ is equal to that of $C_3$. 

The calculation of the output force in Model D is the same as that in Model C, using Equation (\ref{eqn-st8}). 

We used an iterative method to solve the nonlinear algebraic equations above. The output force $F$ can be obtained as a function of the contraction ratio, and the initial conditions with zero contraction are $S_1=L_{10}$ and $S_2=L_{20}$, where the linear IN-FOAM is at the initial length $S_{sum,0}$. Starting with the initial value of $\theta_2$, we make $\theta_2$ increase 0.01 rad in each iteration step, until {\it{F}} drops to zero.

At the beginning of the iteration, Model A is used to initialize the calculation. After each step, whether the model should be changed is judged by the following conditions. When $\theta_1\leq\theta_2$, Model B is adopted, which means $C_2$ starts to contact with $C_3$. If the height of $C_2$ is larger than that of $C_3$, Model A and Model B are respectively changed to Model C and Model D, which means the skin between the skeleton columns comes into contact. When $\theta_1\leq\theta_2$ in Model C, Model D is used for the next step. 

It should be noted that, at the ends of the pouch, the actual inflated shape deviates from the planar assumption (the width is large enough) above, where the cross-section flattens at the heat-sealed edges. When the pouch's aspect ratio ($W/L$) is small, the assumption no longer holds. The linear IN-FOAM consists of pouches similar to the Peano muscle, for which the aspect ratio should be at least 3:1 to minimize the cross-sectional distortion caused by the ends\cite{veale2016characterizing}. Thus, for the linear IN-FOAM, we estimate that this end effect becomes significant when the pouch aspect ratio is lower than 3:1, and the planar assumption is invalid. In our future work, we will investigate the detailed shape of the pouch's ends, to develop a more sophisticated model based on the principle of virtual work. 

From Model C and D we know that there is a residual length $w_2$ between skeleton columns (Fig. \ref{figModel}(c) and (d)), which has no contribution to the contraction ratio and leads to a decrease in the output force ($H=0$ in Equation (15)). To avoid that case, $L_{20}$ needs to be limited with a given $L_{10}$. Here, we provide a rough estimation of the maximum $L_{20}$ by neglecting the compression between the skin and skeleton, as well as the elongation of sheet materials, where the fully inflated pouch has a circular cross-section. We use a V-shaped profile to represent the geometry of the skin between skeleton columns. When the skeleton pouches are fully inflated and the skeleton columns are pulled together (with no distance) by the tensioned skin, to avoid the contact of the top and bottom skin, the length of the V-shaped profile should be no more than the diameter of an inflated skeleton pouch,
\begin{equation}\label{eqn}
L_{20}\leq\frac{2L_{10}}{\pi}.
\end{equation}

\section{Experiments}
\subsection{Characterizations of Linear IN-FOAMs}
\subsubsection{Static and Quasi-static Properties}
We tested the static and quasi-static properties of the linear IN-FOAM on a universal test machine (INSTRON, 68TM-10). The tested sample was made from TPU-coated nylon fabrics with thicknesses of 0.09 mm (skeleton) and 0.17 mm (skin). The sample has a two-layer skeleton, and there are seven columns of pouch units. The sizes of each skeleton unit are $L_{10}=20\;\rm{mm}$, $L_{20}=10\;\rm{mm}$, $W=80\;\rm{mm}$.

During the tests, both ends of the linear IN-FOAM were clamped on the test machine. To measure the blocked force, the linear IN-FOAM was fixed at its initial length and pre-loaded to 1 N. The blocked force was recorded by the test machine. The actuation pressure is regulated by an air supply system (Appendix \ref{appPressureControl}, Fig. \ref{figControl}), measured by air pressure sensors (MPX5100DP, NXP Semiconductors, B.V.), and recorded by a microcontroller board (Arduino Mega 2560 Rev3). To test the force-contraction relation, the linear IN-FOAM was assigned with a contractile and a following recovery displacement at 1 $\rm{mm\cdot s^{-1}}$, which was controlled by the test machine.

\begin{figure*}
    \centering
    \includegraphics[width=1\textwidth]{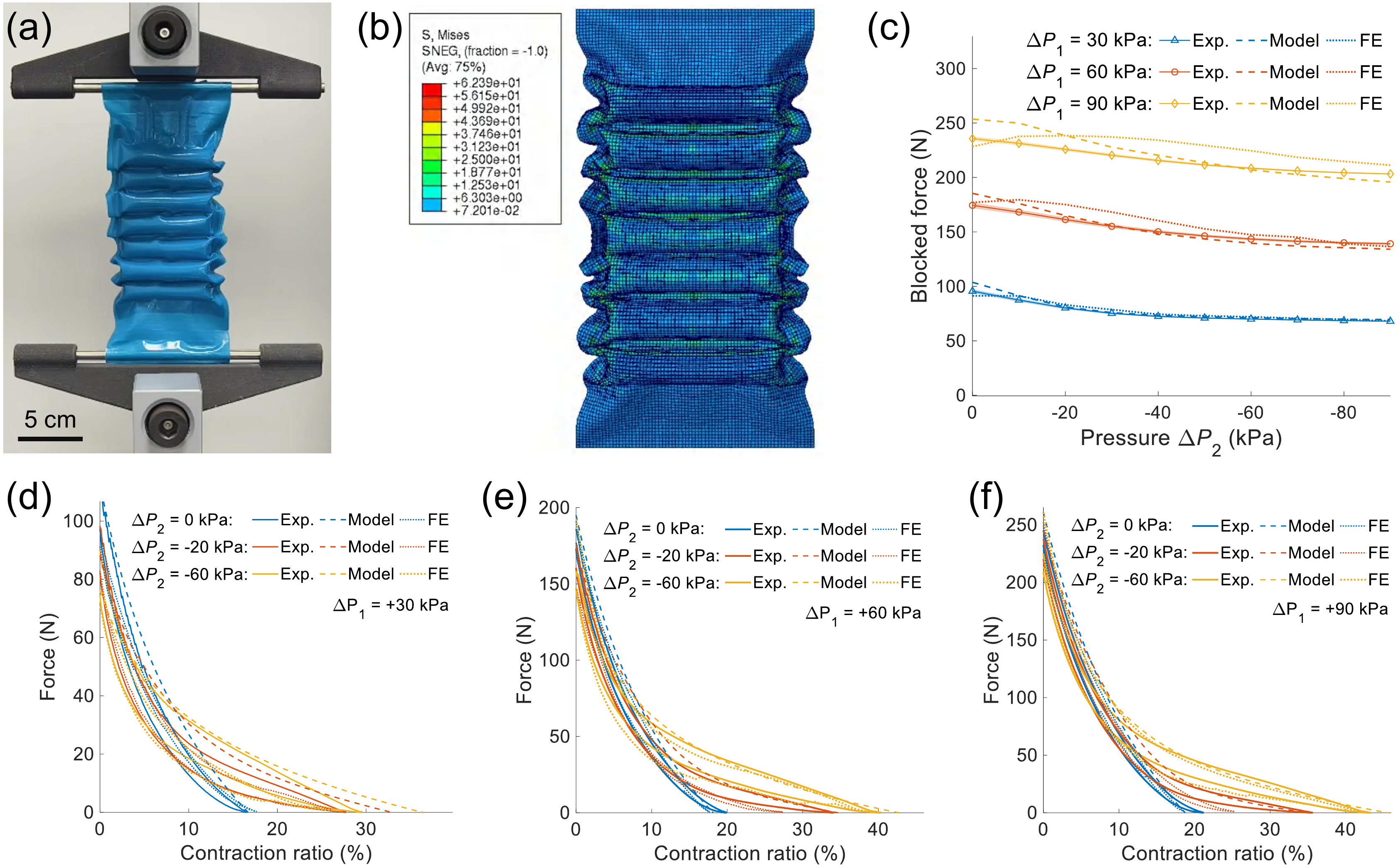}
    
    \caption{Static property of the linear IN-FOAM with constant positive and negative pressures. (c-f) Experimental, theoretical, and simulation results of blocked force and force-contraction. Exp.: experiment; FE: finite element simulation. In the theoretical model, the elastic modulus $E$ is set to 400 MPa, which is roughly in accord with the material's tensile test results. The factor $\delta$ (4\:\%) is estimated by measuring the change in a sample's length before and after the heat-pressing process. (a) The tested sample (size: $31.5\times10.6\times0.08 \:\rm{cm^3}$, in the initial state; weight: 24.6 g; skeleton length: 20 cm) at the maximum contraction (~ 43\:\%, when $\Delta P_1=+90$ kPa, $\Delta P_2=-60$ kPa). (b) The simulated deformation at the maximum contraction, when $\Delta P_1=+90$ kPa, $\Delta P_2=-60$ kPa. The contour depicts von Mises stress (S. Mises), which concentrates on the skeleton's edges. (c) Blocked force as a function of the negative pressure $\Delta P_2$ under different positive pressures $\Delta P_1$. (d-f) Force-contraction relations when $\Delta P_1$ equals +30 kPa, +60 kPa, and +90 kPa, respectively.}

    \label{figStatic1}
\end{figure*}

The blocked force and force-contraction relation of the linear IN-FOAM were tested under two conditions. Condition 1 was keeping the positive and negative pressures constant throughout the experiments. In Condition 2, the skeleton was first inflated and then closed, followed by a constant negative pressure applied to the voids (Fig. \ref{figStatic2}(a)).

Under Condition 1, we measured the blocked force as a function of the applied pressures. The relations are shown in Fig. \ref{figStatic1}(b). The curves show that the blocked force slightly decreases with an increasing negative pressure $\Delta P_2$, and becomes greater with a higher positive pressure $\Delta P_1$. As more air moves into the skeleton, the pressure-induced tension increases, making the output force larger, like most of the pneumatic artificial muscles driven by positive pressure. 
As the negative pressure increases, the output force decreases, which is different from FOAMs\cite{li2017fluid}. The negative pressure would tightly press the skin onto the skeleton, reducing the inflated height of the skeleton, and the blocked force is thus decreased. For the blocked force, the comparison between the theoretical model, FE simulations, and the experimental results is shown in Fig. \ref{figStatic1}(c). The differences probably come from the linear elasticity approximation of the material, as the actual viscoelasticity and hyperelasticity are not considered for simplicity. In addition, the skin can be slightly longer than the skeleton after the heat-sealing process, in which they are bonded at the two ends. This redundant length of the skin would not transmit the tension if it were not fully stretched, so there would be more discrepancies when ${\Delta} P_2$ is low.

The force-contraction relations under Condition 1 are shown in Fig. \ref{figStatic1}(d-f). The output force is a monotonically decreasing function of the contraction ratio. When the magnitude of $\Delta P_2$ increases, the maximum contraction ratio increases, and the curve shows a flattening trend, which means a decrease in the overall stiffness during the contraction. The whole curve rises when $\Delta P_1$ increases, which is because a higher positive pressure generates a higher output force to overcome the contraction resistance, $F_r$, resulting in a higher contraction ratio. The modeling and simulation results demonstrate consistency with the experimental data, and the simulations can accurately predict the detailed deformation during the contraction, including those local wrinkles occurring at the sides (Fig. \ref{figStatic1}(a) and (b)). However, the FE simulation results exhibit slight oscillations when these wrinkles suddenly occur due to local buckling (Appendix \ref{appSimulation}).

\begin{figure}
    \centering
    \includegraphics[width=0.485\textwidth]{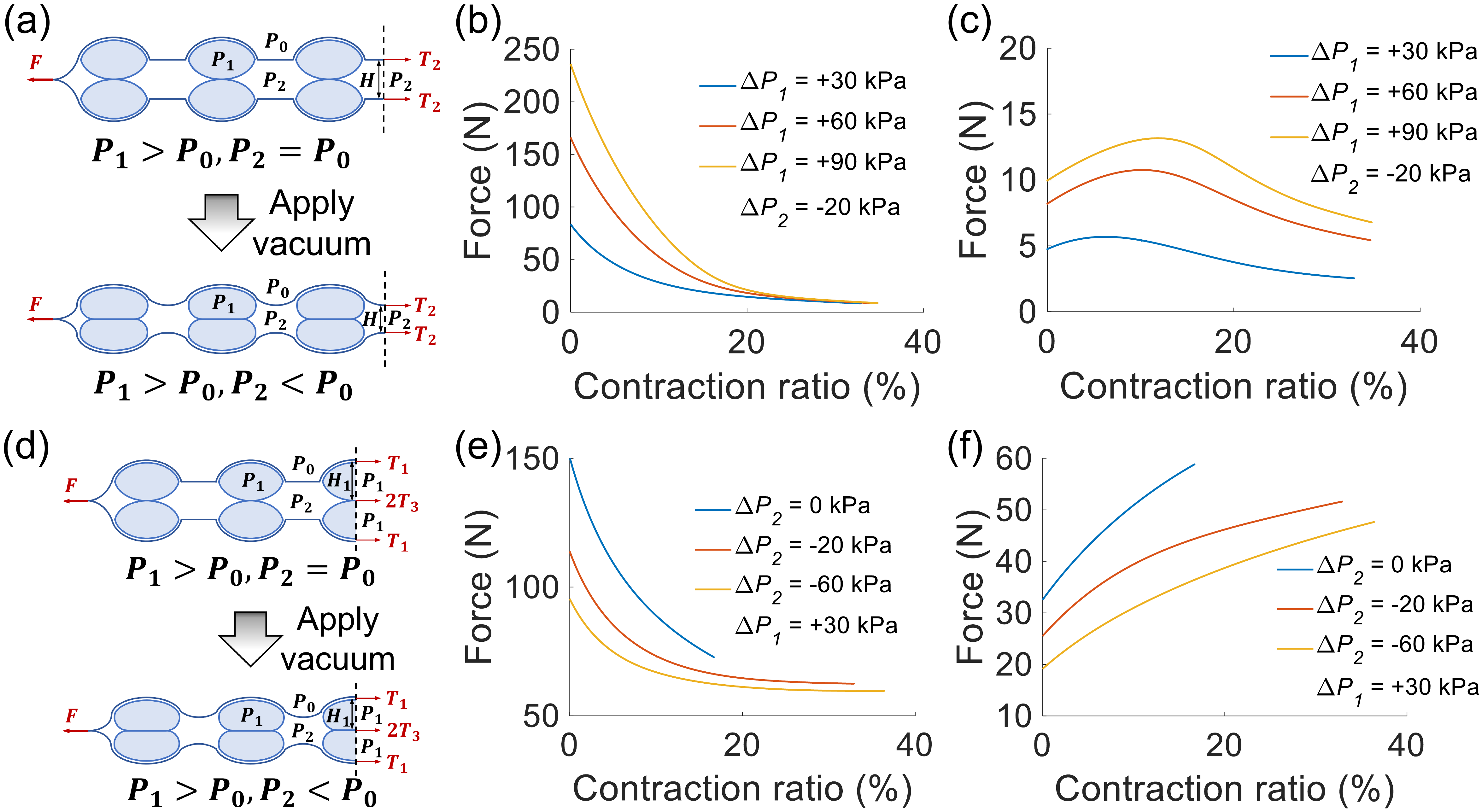}
    \caption{Two analytical approaches to the output force. (a) Approach I: cut the actuator along the center line of the void. The pushing force generated by $(P_0-P_2)$ functions as the drive force for contraction. (b) Tension force $2T_2$ as a function of contraction ratio. (c) Pushing force $(P_0-P_2)HW$ as a function of contraction ratio. (d) Approach II: cut the actuator along the center line of the skeleton pouch. The pushing force generated by $(P_1-P_0)$ functions as resistance to contraction. (e) Tension force $2T_1+2T_3$ as a function of contraction ratio. (f) Pushing force $2(P_1-P_0)H_1W$ as a function of contraction ratio.}
    \label{figPressureEffect}
\end{figure}

The opposite effect of positive pressure and negative pressure on the output force can be explained by our analytical model. According to Equation (15) and Fig. \ref{figPressureEffect}(a), the output force of the linear IN-FOAM mainly consists of the tension of the skin and the pushing force generated by the pressure difference $(P_0-P_2)$. As the positive pressure increases, the skeleton expands and increases the height $H$, making both tension and pushing forces higher (Fig. \ref{figPressureEffect}(b) and (c)), so the output force increases. For the effect of negative pressure, the output force can be analyzed from another perspective. As shown in Fig. \ref{figPressureEffect}(d), the actuator is cut along the center line of a skeleton pouch, and the output force can be expressed as
\begin{equation}\label{eqn}
F=2T_1+2T_3-2(P_1-P_0 ) H_1 W-F_r,
\end{equation}
where $H_1$ is the height of a single skeleton pouch. Now the pushing force generated by $(P_1-P_0)$ functions as resistance to the contraction. We calculate the tension force $(2T_1+2T_3)$ and the pushing force $2(P_1-P_0 ) H_1 W$ under different negative pressures as a function of the contraction ratio (Fig. \ref{figPressureEffect}(e) and (f)). The results show that the tension force and the pushing force both decrease as the negative pressure increases. However, the influence of negative pressure on the tension force differs from that on the pushing force. At lower contraction, the tension force decreases more than the pushing force, so the output force decreases; at higher contraction, the tension force decreases less, so the output force increases. That's why the force-contraction curves under different negative pressures intersect, and a higher negative pressure results in a larger contraction ratio but a lower blocked force.

The relations between the blocked force and $\Delta P_2$ under Condition 2 are significantly different from those under Condition 1. The blocked force and $\Delta P_1$ increase as the magnitude of $\Delta P_2$ increases (Fig. \ref{figStatic2}(b) and (c)). 
Compared to Condition 1 with constant $\Delta P_1$, the closed skeleton can resist the skin's press through the increase of $\Delta P_1$ (Fig. \ref{figStatic2}(c)) due to a decrease in volume of a constant amount of air. Similar to Condition 1, when we increase the initial value of $\Delta P_1$ before applying vacuum to the skin, the output force and the maximum contraction ratio would both become larger (Fig. \ref{figStatic2}(d-f)).

Based on these results, we can tune the static and quasi-static performances of the linear IN-FOAM in multiple ways. This is enabled by the hybrid drive of the positive and negative pressures. The output force is mainly affected by $\Delta P_1$, while the major contribution of $\Delta P_2$ is to increase the maximum contraction further. The operation conditions of the skeleton's chambers, keeping constant $\Delta P_1$ or being closed, can affect the outputs under $\Delta P_2$ through their interactions with the skin. Furthermore, the linear IN-FOAM allows numerous operation modes when taking different airflow conditions, pressurizing sequences, and chamber connectivity into consideration (Appendix \ref{appMultipleOperation}, Fig. \ref{figOperations}). These operation modes enable multiple options for a single actuator and thus can provide enhanced mechanical functionalities compared to those actuators driven by a single pressure source.

\begin{figure}
    \centering
    \includegraphics[width=0.485\textwidth]{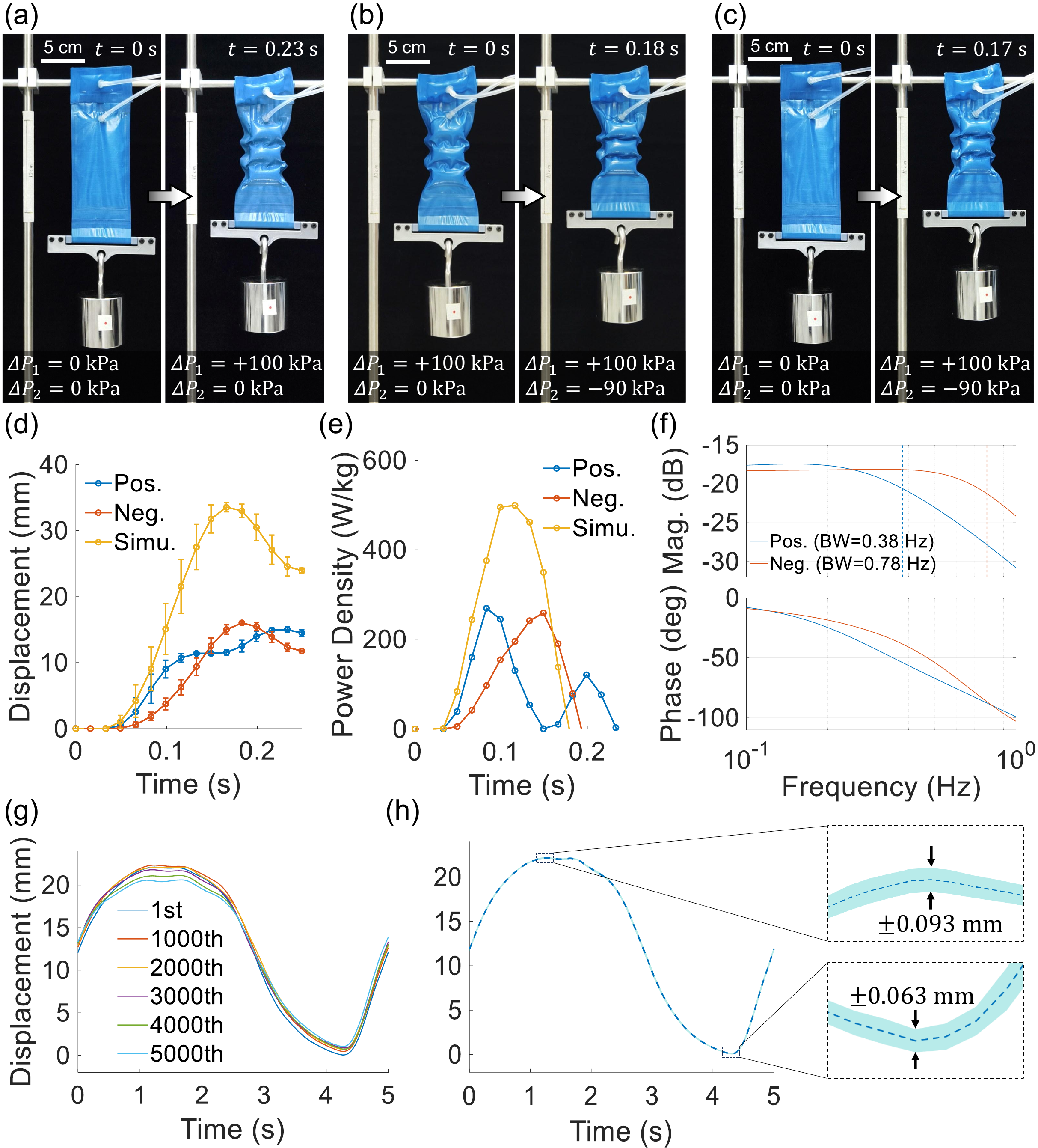}
    
    \caption{Dynamic properties of the linear IN-FOAM. The lightweight actuator (size: $200\times70\times0
    .70 \: \rm{mm^3}$, in the initial state; weight: 7.46 g; skeleton length: 80 mm) is made of TPU-coated nylon fabrics with a thickness of 0.09 mm. The applied pressures were $\Delta P_1 =+100 \:\rm{kPa}$ and $\Delta P_2=-90 \:\rm{kPa}$. (a-c) Actuation process in the power density tests under positive pressure, negative pressure, and simultaneous positive-negative pressure, respectively. (d-f) Dynamic response of the linear IN-FOAM. Pos.: positive-pressure actuation; Neg.: negative-pressure actuation; Simu.: simultaneous positive and negative pressure actuation. (d) Displacement variations in the power density tests. Under the simultaneous actuation, the actuator can lift a 1 kg load for 3.3 cm within 0.2 s. (e) Power density curves during the lifting process. A peak power density of $\sim$ 0.499 kW$\cdot$kg$^{-1}$ and an average power density of $\sim$ 0.272 kW$\cdot$kg$^{-1}$ was achieved in the simultaneous actuation. (f) Mag.: magnitude. Frequency response from target actuation pressure (unit: kPa) to contraction of the actuator (unit: mm). Sinusoidal chirp pressures were generated to drive the actuator. The bandwidths of the linear IN-FOAM under the positive and negative pressure actuation were 0.38 Hz and 0.78 Hz, respectively. (g) Repetitive displacement in the 1st, 1000th, 2000th, 3000th, 4000th, and 5000th cycles (a 1-kg load was applied). (h) Average displacement curve in the first 50 cycles.}

    \label{figDynamic}
\end{figure}

In some aspects of architecture and performances, IN-FOAMs have similarities with pouch motors\cite{niiyama2015pouch} and FOAMs\cite{li2017fluid}. The architecture is similar to FOAMs, including a skin, a skeleton, and fluid media; the maximum force-to-weight ratio ($\sim$ 9.59 kN$\cdot$kg$^{-1}$) is also comparable to that of the linear FOAM ($>$10 kN$\cdot$kg$^{-1}$). The skeleton consists of a series of pouches, and the actuation stress ($\sim$ 108 kPa) is close to that of the linear Pouch Motor ($\sim$ 92 kPa). Since the inflatable skeleton is made from heat-sealable fabrics or films, our artificial muscles are softer than the typical FOAMs and can be compressed to a flat shape and folded when not actuated, making the actuator portable. Combining the contraction of the skin and the skeleton, the maximum contraction ratio of the linear IN-FOAM is approximately 43\:\%, which is higher than that of the linear Pouch Motor ($\sim$ 28\:\%) and the linear Peano Muscle ($\sim$ 17\:\%)\cite{veale2016characterizing}, and slightly lower than that of the FOAM with a spring skeleton ($\sim$ 50\:\%). Besides, the specific force-to-weight ratio, maximum force-to-volume ratio, and specific force-to-volume ratio of the linear IN-FOAM are $0.107\:\rm{kN\cdot kg^{-1}\cdot kPa^{-1}}$, $8,919\:\rm{kN\cdot m^{-3}}$, and $ 99.1\:\rm{kN\cdot m^{-3}\cdot kPa^{-1}}$, respectively. A comparison of IN-FOAMs and relevant pneumatic artificial muscles is provided in Table \ref{tabActuator}.

\subsubsection{Dynamic Properties}
To characterize the dynamic performances (including power density, strain rate, work density, and bandwidth; the calculation formulas of the metrics are provided in Appendix \ref{appMetrics}), we performed a group of load-lifting experiments using a small linear IN-FOAM made from TPU-coated nylon fabric with a 0.09 mm thickness. This actuator has a two-layer skeleton with three pouches in each layer, and its weight is 7.46 g. In the experiments, the artificial muscle was fixed on a stand, loaded with a 1 kg weight. 

For the measurement of power density, strain rate, power-to-volume ratio, and work density (by step response tests, Appendix \ref{appPowerDensity}), a positive pressure of $\Delta P_1=+100$ kPa was applied to the skeleton via an air reservoir, and this case was used as a reference (Fig. \ref{figDynamic}(a)). To test the effects of the additional negative pressure, a step pressure of $\Delta P_2=-90$ kPa was applied to the skin via another air reservoir, while the skeleton was kept at +100 kPa via closed-loop control (Fig. \ref{figDynamic}(b)). In the third case, the pressures of $\Delta P_1=+100$ kPa and $\Delta P_2=-90$ kPa were applied simultaneously (Fig. \ref{figDynamic}(c)). In each case, the linear IN-FOAM was able to lift the load within 0.25 s. The results are shown in Fig. \ref{figDynamic}(d and e). When actuated by positive pressure, negative pressure, and both pressures simultaneously, the peak power densities are 0.269, 0.259, and 0.499 kW$\cdot$kg$^{-1}$, respectively (the weight of the pressure source and the tubing are not included). The maximum power-to-volume ratios are 205, 197, and 380 kW$\cdot$m$^{-3}$, respectively. The peak strain rates are 247, 256, and 475\:\%$\cdot$s$^{-1}$, respectively. The work densities are 14.98, 16.02, and 33.6 kJ$\cdot$m$^{-3}$, respectively. 

\begin{figure*}
    \centering
    \includegraphics[width=1\textwidth]{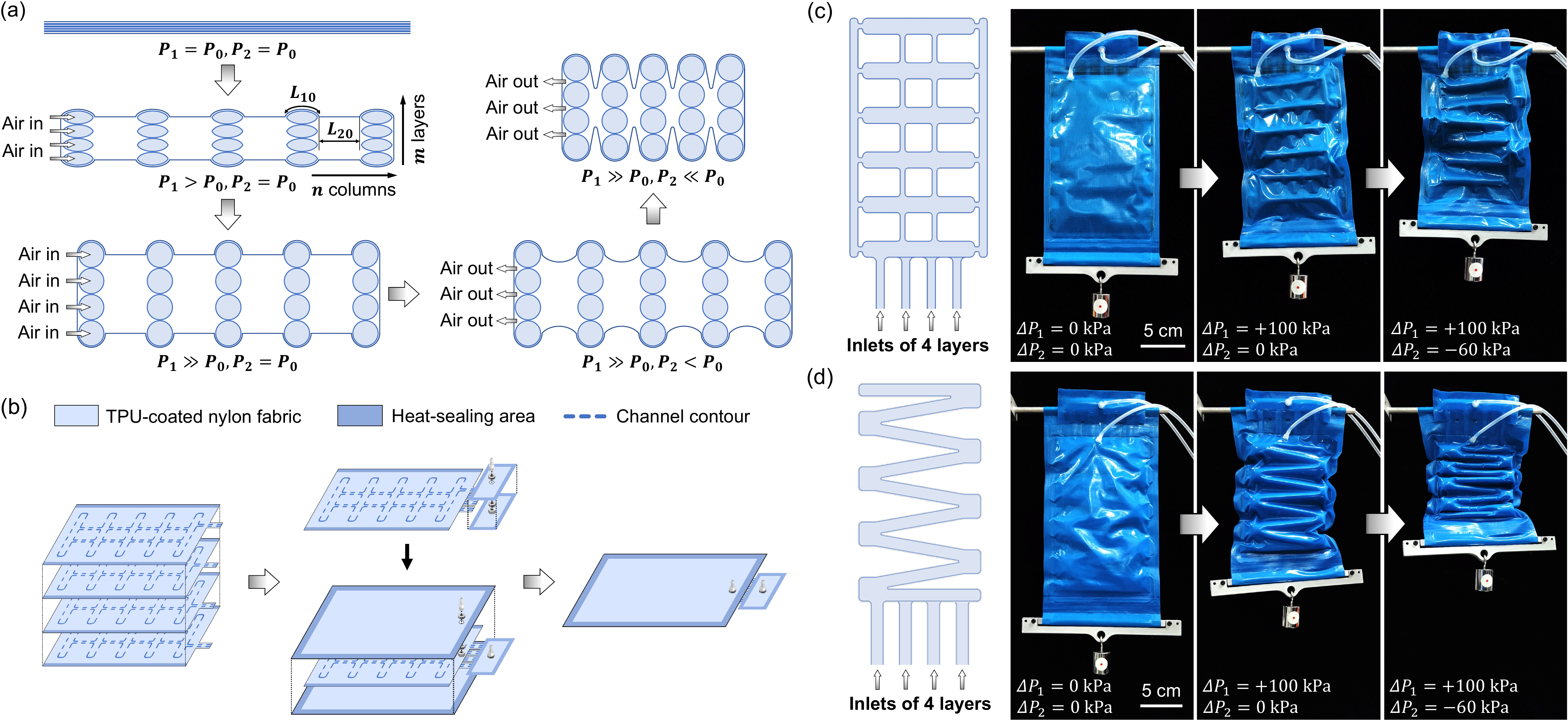}
    
    \caption{Linear IN-FOAMs with multilayer skeletons. (a) A stacked skeleton with multiple layers can increase the maximum contraction ratio. Assuming the initial length of skeleton pouches is fixed, as the inflated height of skeleton columns increases, the length of the skin that can be compressed into the voids becomes larger. Therefore, the maximum contraction ratio increases. (b) Fabrication of linear IN-FOAMs with multilayer skeletons. The preliminary steps are patterning and then heat-pressing the sheet materials, which are shown in Fig. \ref{figPrinciple}(e). (c) A four-layer skeleton with a parallel-pouch pattern has inflated beams on both sides to keep the skeleton columns perpendicular to the contraction direction. With the resistance of the inflated beams, the maximum contraction ratio of the actuator (size: $26.5\times13\times0.2 \:\rm{cm^3}$; weight: 48.1 g; skeleton length: 15 cm) is 26\:\%, consisting of 13\:\% and 13\:\% under the positive and negative pressure actuation, respectively. (d) A four-layer skeleton with a zigzag pattern enables the actuator (size: $30.3\times15\times0.2 \:\rm{cm^3}$; weight: 44.3 g; skeleton length: 18.3 cm) to achieve a maximum contraction ratio of 54\:\%, consisting of 29\:\% and 25\:\% under the positive and negative pressure actuation, respectively.}

    \label{figMultilayer}
\end{figure*}

For the bandwidth test, a sinusoidal chirp positive-pressure was applied to the skeleton, with a frequency from 0.1 to 1 Hz, increased by 0.1 Hz after each cycle. In the second case, the skeleton pressure was first kept at +100 kPa, and another sinusoidal chirp negative-pressure was then applied to the skin. The bandwidths of these two cases were 0.38 Hz and 0.78 Hz, respectively (Fig. \ref{figDynamic}(f)). However, with the tested actuator as load, the bandwidths of our pneumatic system (from the target pressure to the actual pressure) in positive and negative pressure controls were only 0.64 Hz and 0.79 Hz, respectively, which limits the bandwidth of the actuator.

The energy conversion efficiency of the linear IN-FOAM varies along its contraction (Appendix \ref{appEnergy}, Fig. \ref{figEnergy}(e-g)). While the energy efficiencies under the positive and negative pressure actuation over the entire stroke are respectively 5.5\:\% and 8.4\:\%, the peak energy efficiencies are 57\:\% and 22\:\%, respectively. By pre-inflating the skeleton and then closing it, the peak energy efficiency under the negative-pressure actuation increases to 36\:\%, and the average energy efficiency increases to 11.5\:\%. When the linear IN-FOAM is driven by transferring air from the skin to the skeleton, the peak energy efficiency increases to 64\:\%, and the efficiency in the first 70\:\% of the stroke remains above 29\:\%.

These results indicate that the applied negative pressure accelerates the response of the linear IN-FOAM to the pressurization, decreasing the time for the entire contraction from 0.23 s to 0.17 s and increasing the bandwidth from 0.38 Hz to 0.78 Hz. The power density under simultaneous actuation of the positive and negative pressures (0.499 kW$\cdot$kg$^{-1}$) is approximately the sum of the power densities under these two pressures (0.269 kW$\cdot$kg$^{-1}$ and 0.259 kW$\cdot$kg$^{-1}$ under the positive and negative pressure actuation, respectively), exhibiting a superposition property. Similar to the static and quasi-static properties, the dynamic properties present diverse results under different operation modes, providing multiple options for practical use.

\subsubsection{Cyclic Test}
To further evaluate the performance under long-term working conditions, we conducted a cyclic test on the linear IN-FOAM with a 1-kg load. Periodic positive and negative pressures (0 kPa to +100 kPa and 0 kPa to -60 kPa, 0.2 Hz) were applied to the skeleton and skin chambers, respectively. The displacement of the actuator in specific cycles was tracked (Fig. \ref{figDynamic}(g)). The average displacement curve in the first 50 cycles was calculated (Fig. \ref{figDynamic}(h)). The standard deviation of the displacement in the first 50 cycles is within 0.5\:\% of the stroke (±0.093 mm at the peak and ±0.063 at the valley). The displacement curve shows good consistency within the first 1000 cycles, after which the performance starts to decay. The actuator's maximum contraction in the 5000th cycle decreased 10.9\:\% relative to that in the first cycle. We observed the heat-sealing area boundary and the material surface between the two skeleton layers. There was no failure caused by interaction between different layers of material. However, the fluidic channel’s boundary in the skeleton was torn apart by the inflation process over time, which led to the deterioration.



\subsection{IN-FOAMs with Multilayer Skeletons}

Based on the working principle and modeling of the linear IN-FOAM, the maximum contraction ratio can be improved by increasing the cross-sectional area, as shown in Fig. \ref{figMultilayer}(a). When the skeleton is inflated, the skin can be compressed into the voids between the skeleton columns under negative pressure. The maximum skin length that can be compressed into the voids depends on the inflated skeleton's height. A simplified geometric model is developed to explain how the maximum contraction ratio increases with the number of the skeleton's layers.

As shown in Fig. \ref{figMultilayer}(a), the linear IN-FOAM has {$m$} layers of {$n-$}column skeleton. The initial lengths of each skeleton pouch and the skin between the adjacent skeleton columns are $L_{10}$ and $L_{20}$, respectively. A V-shaped profile is used to represent the skin between skeleton columns. Similar to the estimation of the maximum $L_{20}$ in Section III, to avoid the contact of the top and bottom skin and get a complete contraction, when the skeleton columns are fully inflated and the skeleton columns are pulled together with no distance, the length of the V-shaped profile should be no more than the height of (m-1) stacked inflated pouches, $L_{20}\le(m-1) L_{10}\cdot2/\pi$.

\begin{figure}
    \centering
    \includegraphics[width=0.485\textwidth]{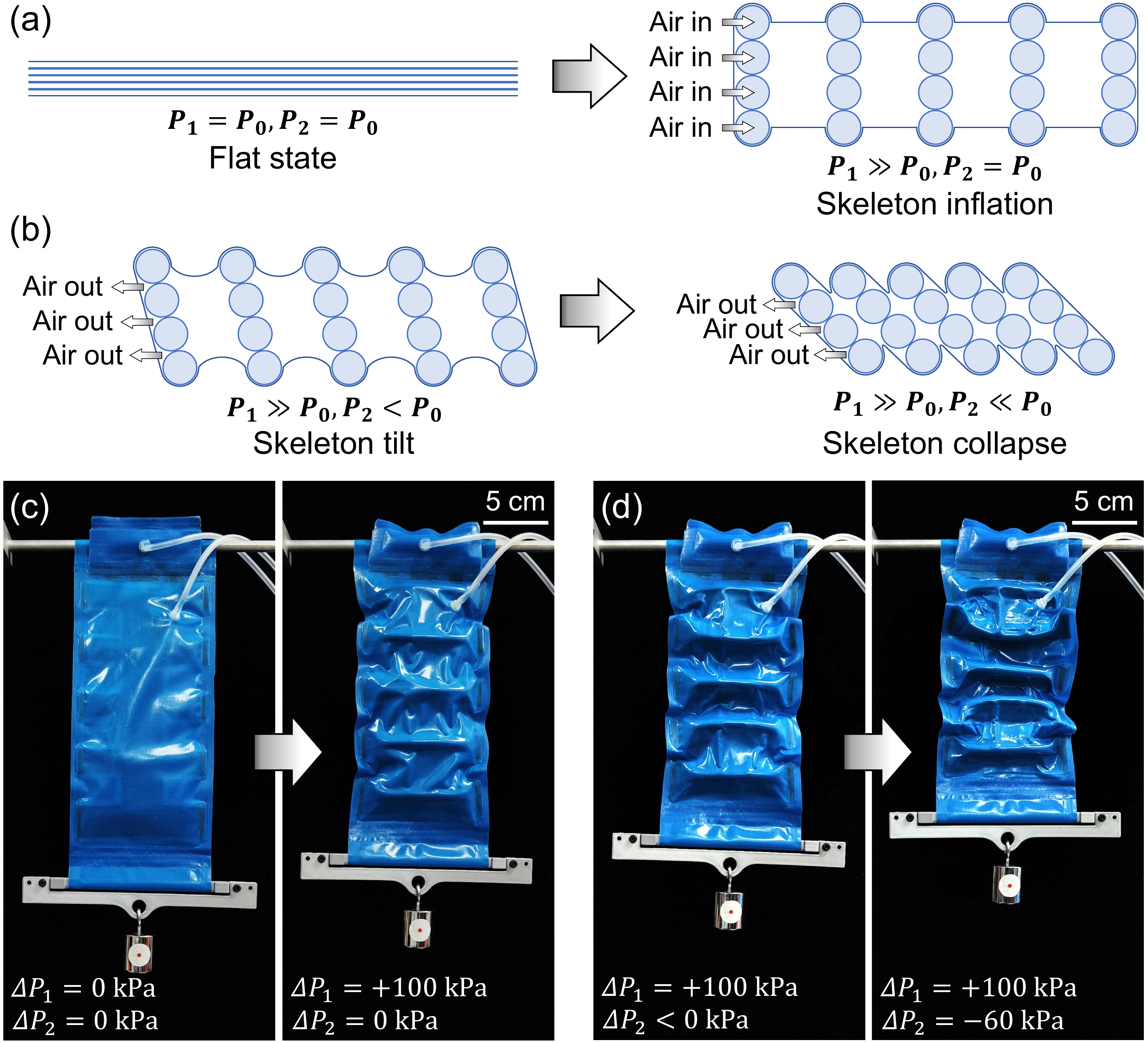}
    
    \caption{Skeleton collapse during the contraction of the IN-FOAM with a multilayer parallel-pouch skeleton. (a) Schematic of skeleton inflation (b) Schematic of skeleton collapse under asymmetric tension in the skin during negative pressure actuation. In the experiment, the skeleton may collapse in different directions, not aligned in order, as this illustration shows. (c) Skeleton inflation of an IN-FOAM with a multilayer parallel-pouch skeleton. (d) Skeleton collapse of the multi-layer IN-FOAM.}

    \label{figCollapse}
\end{figure}

The maximum contraction ratios under the actuation of positive and negative pressures can be calculated as
\begin{equation}\label{eqn}
CR_{+,\:max}=\frac{n(1-2/\pi) L_{10}}{nL_{10}+(n-1) L_{20}}\times100\:\%,
\end{equation}
\begin{equation}\label{eqn}
CR_{-,\:max}=\frac{(n-1) L_{20}}{nL_{10}+(n-1) L_{20}}\times100\:\%,
\end{equation}
and the maximum contraction ratio in total is the sum of those two above, $CR_{max}\it=CR_{+,\:max}+CR_{-,\:max}$.

When $L_{20}\rightarrow0$, the theoretical maximum contraction ratio is equal to that of pouch motors, which is $(\pi-2)/\pi\approx36\:\%$. When $L_{10}\rightarrow0$, IN-FOAMs are similar to FOAMs with skeletons made from thin materials, with the maximum contraction ratio close to 100\:\%. For IN-FOAMs, a larger $L_{20}/L_{10}$ can increase the maximum contraction ratio. When $L_{20}=(m-1) L_{10}\cdot2/\pi$, the maximum contraction ratios under the actuation of positive and negative pressures are
\begin{equation}\label{eqn}
CR_{+,\:max}^*=\frac{(\pi-2)n}{\pi n+2(m-1)(n-1)}\times100\:\%,
\end{equation}
\begin{equation}\label{eqn}
CR_{-,\:max}^*=\frac{2(m-1)(n-1)}{\pi n+2(m-1)(n-1)}\times100\:\%.
\end{equation}

\begin{figure*}[!t]
    \centering
    \includegraphics[width=1\textwidth]{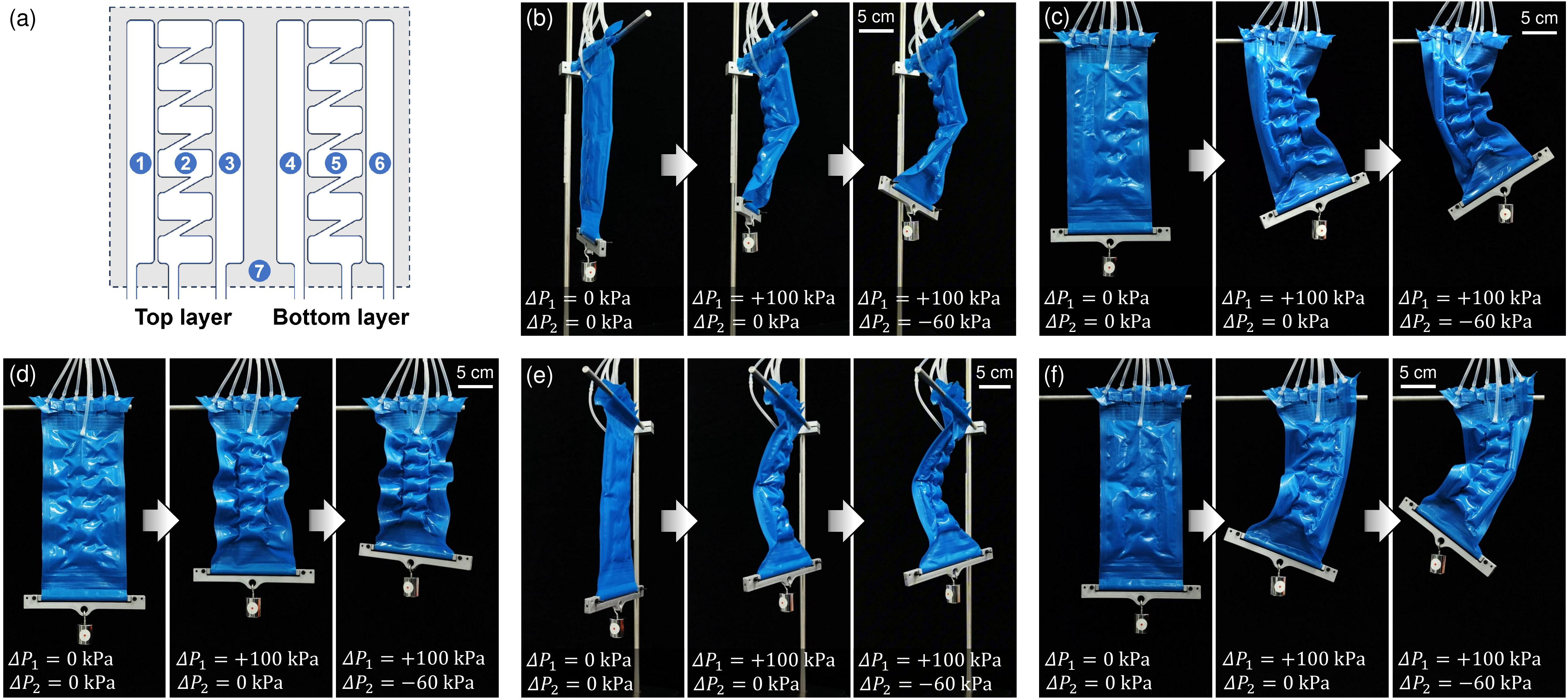}
    
    \caption{A multi-mode IN-FOAM with five motion modes. Motion modes were tested with a 100 g load hung below the IN-FOAM (size: $32\times12\times0.08 \:\rm{cm^3}$; skeleton length: 18 cm; weight: 31.5 g). The actuation pressures were set to $\Delta P_1 =+100 \:\rm{kPa}$ and $\Delta P_2 =-60 \:\rm{kPa}$. (a) Pneumatic circuit of the multi-mode IN-FOAM. There are six independent channels in the two-layer skeleton, which are denoted by Channels 1-6. The void between the skin and the skeleton is denoted by Channel 7. (b) The actuator bends forward with Channels 2, 4, 5, and 6 inflated. Then it bends further after applying vacuum to Channel 7. (c) The actuator bends leftward with Channels 2, 3, 5, and 6 inflated. After applying the vacuum, it bends further. (d) The actuator performs linear contraction with Channels 2 and 5 inflated, and performs a further contraction under the vacuum. (e) The actuator bends backwards with Channels 1, 2, 3, and 5 inflated, and then bends further under the vacuum. (f) The actuator bends rightwards with Channels 1, 2, 4, and 5 inflated, and then bends further under the vacuum.}

    \label{figMultimode}
\end{figure*}

\begin{table}
 \centering
 \caption{Theoretical values of maximum contraction ratio for the linear IN-FOAMs with multilayer skeletons.}
  \begin{tabular}[htbp]{@{}cccccc@{}}
    \hline
     & $n=1$ & $n=2$ & $n=3$ & $n=4$ & $n=5$ \\
    \hline
    $m=2$ & 36.3\:\% & 51.7\:\% & 55.3\:\% & 56.9\:\% & 57.8\:\% \\
    $m=3$ & 36.3\:\% & 61.1\:\% & 65.6\:\% & 67.4\:\% & 68.5\:\% \\
    $m=4$ & 36.3\:\% & 67.4\:\% & 72.0\:\% & 73.8\:\% & 74.8\:\% \\
    $m=5$ & 36.3\:\% & 72.0\:\% & 76.4\:\% & 78.1\:\% & 79.0\:\% \\
    $m=6$ & 36.3\:\% & 75.4\:\% & 79.6\:\% & 81.2\:\% & 82.1\:\% \\
    $m=+\infty$ & 36.3\:\% & $\to100\:\%$ & $\to100\:\%$ & $\to100\:\%$ & $\to100\:\%$ \\
    \hline
  \end{tabular}
  \label{tabMultilayer}
\end{table}

The predicted maximum contraction ratios with different groups of $m$ and $n$ are listed in Table \ref{tabMultilayer}. We fabricated an IN-FOAM with $m=4,\;n=5$ (size: $27\times10.3\times0.2\:\rm{cm^3}$; weight: 35.2 g; skeleton length: 17.4 cm; Fig. \ref{figCollapse}(c)) and measured its maximum contraction ratio (the contraction ratio test was repeated 10 times). The maximum contraction ratio under the positive-pressure actuation was 9.8\:\% and further increased to 27.7\:\% after applying a vacuum. The result is far less than the theoretical prediction from our geometric model. Such a difference is probably due to the unexpected buckling of the skeleton (Fig. \ref{figCollapse}(d)). When it contracts under the negative-pressure actuation, the skeleton columns tend to tilt, fall, and finally collapse (Fig. \ref{figCollapse}(b)), reducing the maximum contraction ratio. This is similar to the unstable interfacial sliding of two compressed balloons.


We modified the architecture of the multilayer skeleton to increase its overall stability. The first solution is to add inflated beams between the skeleton columns, directly preventing the collapse (Fig. \ref{figMultilayer}(c)). The measured maximum contraction ratio is 26\:\% and is still much lower than the theoretical prediction. The inflated beams would resist the contraction and cause the skeleton to twist, and as a consequence, the skeleton could not contract completely. In the second solution, the multilayer skeleton has zigzag-shaped fluidic channels (Fig. \ref{figMultilayer}(d)). The skeleton columns are connected directly, and the structural stiffness to prevent collapse is reinforced. The skeleton can be flexibly folded up under the skin's tension with a reduced contraction resistance compared to the previous designs. The maximum contraction ratio reaches 54\:\%, though there is still little space between skeleton columns at the final contraction state. The former geometric model can be modified for the zigzag skeleton:
\begin{equation}\label{eqn}
CR_{max}^{'}=\frac{S_{sum,0}-nL_{10}\cdot2/\pi}{S_{sum,0}}\times100\:\%,
\end{equation}
where $n$ is the number of zigzag edges (including the horizontal channel at both ends). For the zigzag skeleton shown in Fig. \ref{figMultilayer}(d), $S_{sum,0}=200\;\rm{mm}$, $n=8$, $L_{10}=10\;\rm{mm}$, and the theoretical maximum contraction ratio is $CR_{max}^{'}=74.5\:\%$. The difference between experimental and theoretical values may be further reduced by increasing the layer number, so that the remaining space for the contraction can be utilized. Taking advantage of these multilayer zigzag skeletons, the IN-FOAMs can realize a higher contraction ratio ($>$43\:\%).

Here, we use the linear IN-FOAM as an example to demonstrate that a high contraction ratio can be achieved by employing a multilayer skeleton. The method can be extended to motions mainly happening in the skeleton plane (e.g., leftward/rightward bending and rotation). For these motions, multilayer skeletons can deform consistently in parallel planes to increase the deformation ratio. However, other motions like upward/downward bending and twisting cause significant deformation out of the skeleton plane. For multilayer skeletons of these motions, with both sides of all layers sealed, the outer layers and inner layers cannot deform consistently and generate resistance to prevent the intended deformation.

\begin{figure*}
    \centering
    \includegraphics[width=1\textwidth]{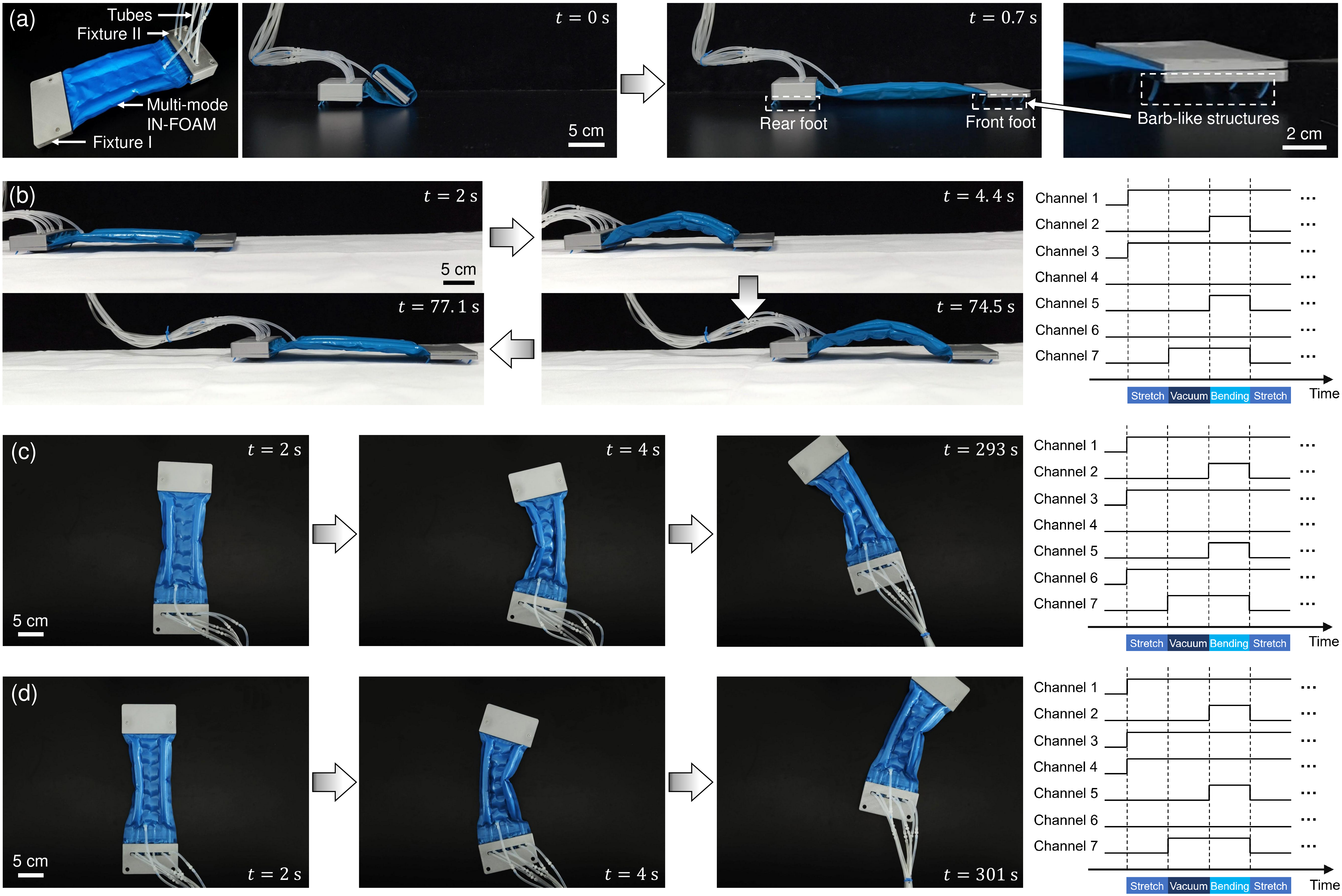}
    \caption{A crawling robot actuated by a multi-mode IN-FOAM. The actuator's pneumatic circuits and multi-mode motions are shown in Fig. \ref{figMultimode}. (a) Structure of the robot (size: $23\times12\times3.2 \:\rm{cm^3}$; weight: 144 g). Both ends of the actuator are clamped in 3D-printed fixtures. Barbed feet are stuck under the fixtures to achieve unidirectional resistance during the robot's motions. When unactuated, the IN-FOAM is flat, and the robot can be folded up for portability. In the working process, the robot can be unfolded by skeleton inflation. (b-d) The robot moves forward, turns left, and turns right on flat surfaces. Corresponding pressure sequences in different channels follow snapshots of the robot's motions.}
    \label{figCrawling}
\end{figure*}

\subsection{Programmable Motions}

We've developed a theoretical model for the linear IN-FOAM in Section III, which explains the working principle of hybrid positive and negative pressure actuation and guides the skeleton design. However, it is not applicable to IN-FOAMs designed for other motions, which have more complex geometries. Here, taking the actuator with a flasher-origami-like skeleton as an example, we demonstrate that FE simulation can be used to predict the deformation of IN-FOAMs with complex skeleton patterns, which provides a foundation of computational design for programmable motions.

Simulation results of the rotary IN-FOAM are shown in Fig. \ref{figSimulation}(g-i). The skeleton is inflated and rotates a small angle under positive pressure actuation, followed by a further rotation under negative pressure actuation, which is similar to the experimental results (Fig. \ref{figMotions}(d)). Besides, when the outer part of the skeleton rotates relative to the inner part, wrinkling occurs between the segments of the fluidic channel (Fig. \ref{figMotions}(d)). This can also be predicted numerically (Fig. \ref{figSimulation}(h) and (i)). In summary, the FE simulation can capture both the overall motions and the local deformations. The details of simulation setups are provided in Appendix \ref{appSimulation}.

\subsection{Integration of Multi-mode Motions}
Multi-mode motions can be produced by directly combining typical artificial muscles with single-mode motions. However, the combined structure is complicated, bulky, and difficult to fabricate. For IN-FOAMs, multi-mode motions can be achieved by introducing a multi-channel skeleton, whose fabrication is similar to the aforementioned processes. When different channels are inflated, the IN-FOAM will perform different motions during actuation. Consequently, multi-mode motion can be integrated into a single IN-FOAM with a compact structure.

A two-layer skeleton with six channels was designed to demonstrate the integration of multi-mode motions. Five motion modes are included in the actuator. As shown in Fig. \ref{figMultimode}(a), Channels 2 and 5 are a series of pouches while Channels 1, 3, 4, and 6 are inflatable beams. When Channels 2 and 5 are inflated, the actuator can perform a linear contraction (Fig. \ref{figMultimode}(d)). The actuator can perform a forward bending with Channels 2, 4, 5, and 6 inflated (Fig. \ref{figMultimode}(b)) and a backward bending with Channels 1, 2, 3, and 5 inflated (Fig. \ref{figMultimode}(e)). The actuator can also bend leftward with Channels 2, 3, 5, and 6 inflated (Fig. \ref{figMultimode}(c)) and bend rightward with Channels 1, 2, 4, and 5 inflated (Fig. \ref{figMultimode}(f)). When a negative pressure is applied, all of these motions can be enhanced.

\begin{figure*}
    \centering
    \includegraphics[width=1\textwidth]{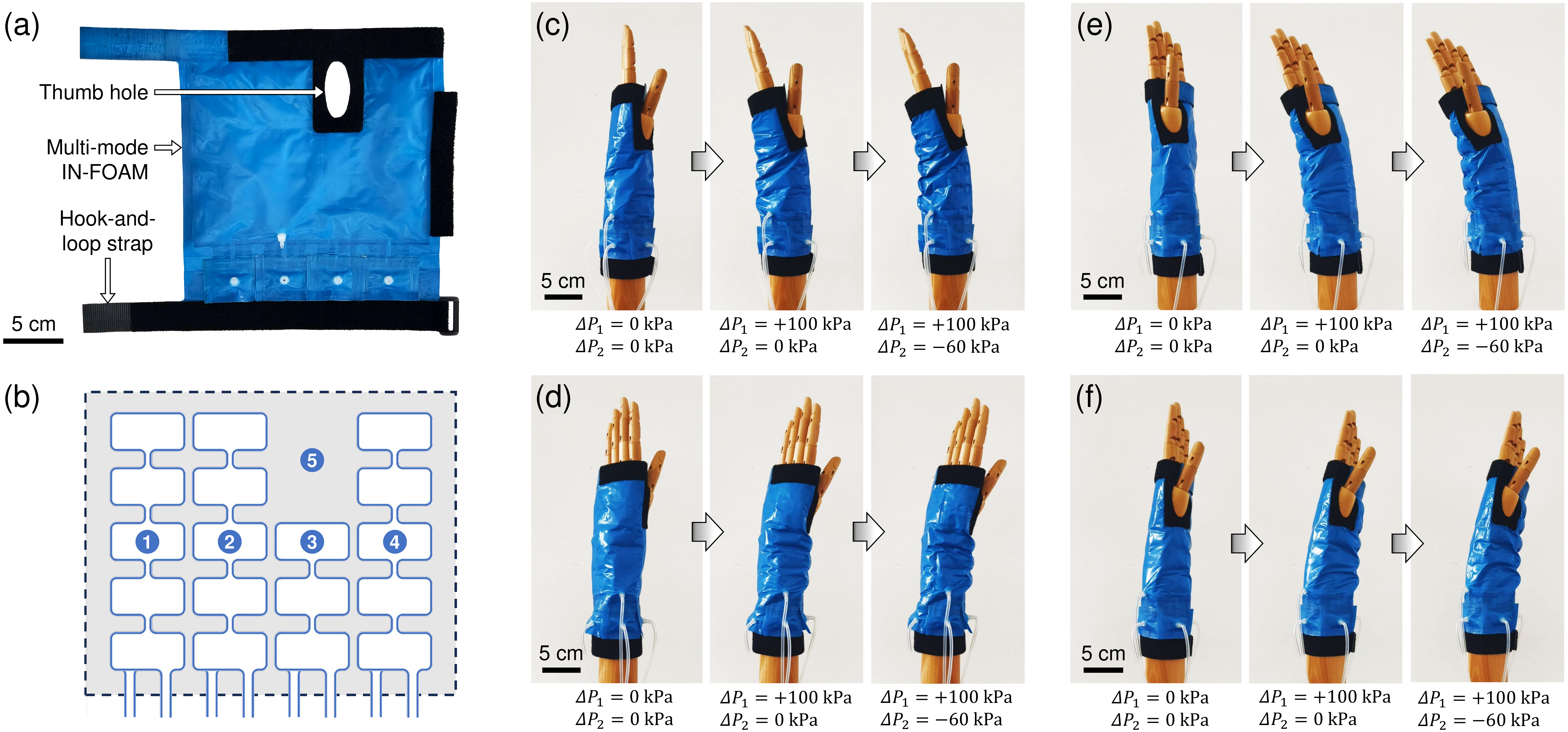}
    \caption{A soft haptic sleeve actuated by a multi-mode IN-FOAM. (a) Structure of the sleeve (main-body size: $250\times215\times0.65 \:\rm{mm^3}$; weight: 51.5 g). The sleeve is designed to be fastened around the wrist via hook-and-loop straps. (b) Pneumatic circuit of the multi-mode IN-FOAM employed in the sleeve. (c-f) The sleeve can convey multiple kinesthetic cues to the wrist by inflating different channels: wrist extension (c); wrist radial deviation (d); wrist extension with radial deviation (e); wrist flexion (f).}
    \label{figWearable}
\end{figure*}

\section{Demonstration}
\subsection{Application in Soft Robotics}

Using the proposed IN-FOAM with five motion modes, we built a soft crawling robot to demonstrate the application of IN-FOAMs. As shown in Fig. \ref{figCrawling}(a), the feet with barb-like structures are affixed to fixtures on both ends of the multi-mode IN-FOAM. With an anisotropic arrangement, the feet can resist movements towards the ends of the barb-like structure. 

By inflating Channels 1, 2, 3, and 5, the robot can bend its body, with the rear foot moving towards the front foot. After further deflating Channels 2 and 5, the robot recovers to a straight state, with the front foot moving away from the rear foot. The robot can move forward like an inchworm using this gait (Fig. \ref{figCrawling}(b)). When the multi-mode IN-FOAM bends leftward, the front foot and rear foot are respectively oriented to the left front and right front. After the robot straightens by deflating Channels 2 and 5, the front and rear feet move a distance along their respective orientations, making the robot rotate a certain angle counterclockwise (Fig. \ref{figCrawling}(c)). By repeating the bending-straightening process, the robot can gradually turn left. Similarly, the robot can turn right when the actuator bends rightward and straightens repeatedly(Fig. \ref{figCrawling}(d)). To prevent the inflated beams from unexpectedly buckling, we applied a vacuum to Channel 7 before activating the bending, using the layer jamming mechanism to increase the body's stiffness.

\subsection{Application in Soft Wearables}
IN-FOAMs can generate multiple motions with a low-profile initial state, which could be used for haptic interaction and rehabilitation in the future. A soft haptic sleeve was built using a multi-mode IN-FOAM. As shown in Fig. \ref{figWearable}(a), hook-and-loop strips were added to the multi-mode IN-FOAM as attachments. The sleeve is made of fabric, enabling seamless integration with clothing. Compared to other pneumatic haptic sleeves, IN-FOAMs enable multiple kinesthetic cues in a low-profile form (Table \ref{tabWearable}).

By inflating different fluidic channels (Fig. \ref{figWearable}(b)), the sleeve can exert indentation and convey kinesthetic cues to the user. The sleeve can provide kinesthetic cues for wrist extension with Channel 1 inflated (Fig. \ref{figWearable}(c)), wrist radial deviation with Channel 3 inflated (Fig. \ref{figWearable}(d)), and wrist flexion with Channel 4 inflated (Fig. \ref{figWearable}(f)). The sleeve can also deliver a kinesthetic cue for wrist extension with radial deviation with Channel 2 inflated (Fig. \ref{figWearable}(e)). When a negative pressure is applied, the kinesthetic cues can be enhanced.

\section{Conclusion}
In this work, we propose IN-FOAMs, which are pneumatic artificial muscles made of heat-sealable sheet materials and driven by a hybrid positive-negative air pressure. IN-FOAMs are thin and portable when not actuated, and they can achieve a significant maximum contraction ratio (\textgreater40\:\%), surpassing conventional positive pressure-driven fluidic artificial muscles. Compared to traditional artificial muscles with a single input, the performance of IN-FOAMs can be more flexibly programmed by adjusting actuation pressures and operation modes. Furthermore, the outputs of force and contraction can be enhanced using multilayer skeletons. Additionally, IN-FOAMs are capable of producing a variety of motions with programmed skeleton patterns. A multi-channel skeleton design allows for integrating multiple motion modes within a single IN-FOAM. 

\setcounter{figure}{0}
\setcounter{equation}{0}
\setcounter{table}{0}
\renewcommand{\theequation}{A\arabic{equation}}
\renewcommand{\thefigure}{A\arabic{figure}}
\renewcommand{\thetable}{A\Roman{table}}

{
\renewcommand{\arraystretch}{4.5}
\begin{table*}[!b]
 \centering
 \scriptsize
 \begin{threeparttable}
 \caption{Comparison of pneumatic artificial muscles.}
  \begin{tabular}[!b]{@{}ccccccccc@{}}
    \hline
    Reference & \makecell{Low-profile\\(initial state)} & Flexibility & \makecell{Motion\\programmability} & \makecell{Maximum\\contraction\\ratio\\(\%)} & \makecell{Maximum\\actuation\\stress\\(MPa)} & \makecell{Maximum\\force-to-weight\\ratio\\($\rm{kN\cdot kg^{-1}}$)} & \makecell{Specific\\force-to-weight\\ratio\\($\rm {kN\cdot kg^{-1}\cdot kPa^{-1}}$)} & Operation pressure \\ 
    \hline
    \makecell{IN-FOAMs\\(this work)} & Yes & High & \makecell{1D/2D contraction\\Bending\\Twisting\\Rotation} & $>$54\:\% & \makecell{0.108\\(+100 kPa\\-10 kPa)} & $>$9.59 & \makecell{0.107\\(+90 kPa\\-0 kPa)} & \makecell{0 $\sim$ +100 kPa\\0 $\sim$ -100 kPa} \\
    FOAMs\cite{li2017fluid} & No & Low & \makecell{1D/2D/3D contraction\\Bending\\Twisting\\Rotation} & $>$90\:\% & \makecell{0.64\\(-90 kPa)} & $>$10 & \makecell{0.1*\\(-100 kPa)} & 0 $\sim$ -100 kPa \\
    \makecell{Pouch\\Motors\cite{niiyama2015pouch}} & Yes & High & \makecell{1D contraction\\Bending\\Rotation} & 28\:\% & \makecell{0.092*\\(+40 kPa)} & Unknown & Unknown & 0 $\sim$ +40 kPa \\
    \makecell{Pleated\\PAMs\cite{villegas2012third}} & No & High & \makecell{1D contraction} & 38.2\:\% & \makecell{0.618\\(+400 kPa)} & 50.2 & \makecell{0.126*\\(+400 kPa)} & +2 $\sim$ +400 kPa \\
    \makecell{McKibben\\Muscles\cite{chou1996measurement,daerden2002pneumatic}} & No & High & \makecell{1D contraction} & 30\:\% & \makecell{1.3\\(+500 kPa)} & 13* & \makecell{0.043*\\(+300 kPa)} & +100 $\sim$ +500 kPa \\
    APAMs\cite{usevitch2018apam} & No & Low & \makecell{1D contraction\\Bending} & 93.1\:\%* & \makecell{0.035*\\(-20 kPa\\+13.79 kPa)} & Unknown & Unknown & \makecell{-20 $\sim$ +10.3 kPa\\0 $\sim$ +20 kPa} \\
    \makecell{PNP\\linear\\brake\cite{jang2022positive}} & No & High & \makecell{1D contraction} & 27.3\:\% & Unknown & 2.24* & \makecell{0.045*\\(-25 kPa\\+25 kPa)} & \makecell{0 $\sim$ -100 kPa\\0 $\sim$ +30 kPa} \\
    \makecell{Hyper-VAMs\cite{coutinho2023hyperbaric}} & No & Low & \makecell{1D contraction} & 80.2\:\% & \makecell{0.0389*\\(-40kPa\\+0 kPa)} & 2.47* & \makecell{0.062*\\(-40 kPa\\+0 kPa)} & \makecell{0 $\sim$ -100 kPa\\0 $\sim$ +60 kPa} \\
    \hline
  \end{tabular}
  \label{tabActuator}
  
  \begin{tablenotes}[flushleft]
    \footnotesize
    \item The operation pressures for pneumatic artificial muscles using simultaneous positive and negative pressures are listed in the order of the inner chamber and the outer chamber.
        
    \item *These data are not directly given by the corresponding references. We use the original data or read the original data from graphs to calculate the properties.
  \end{tablenotes}
  \end{threeparttable}
\end{table*}
}

Despite these benefits, the current version of IN-FOAMs has some limitations. For instance, the buckling or wrinkles of the sheet material can resist contraction and cause unintended motions, negatively affecting the performance of IN-FOAMs. Moreover, the contraction ratio of multilayer IN-FOAMs is still lower than expected,  as the skin tends to compress the stacked skeleton layers. In our future work, the sheet material's stiffness associated with the wrinkles will be investigated to optimize the skeleton patterns. The arrangement and control of multiple channels will also be improved to achieve smoother multi-mode motions.



\appendices
\section{Calculation of mechanical properties}
\label{appMetrics}
The symbols are defined as follows: $x$-displacement of the actuator, $L_{initial}$-initial length of the actuator's skeleton, $F$-output force, $A$-maximum cross-sectional area during actuation, $m$-mass of the actuator, $V_{flat}$-volume of the actuator in the initial flat state, $\Delta P$-pressure difference, $m_L$-mass of load, $h$-height increase of the load lifted by the actuator, $t$-test duration, $\delta t$-time interval between adjacent frames of the video used to record the actuator's contraction, $\delta x$-the actuator's displacement within a time interval, $\delta h$-height increase of the load within a time interval,$g$-gravitational acceleration, $q$-volumetric flow rate. The characterization uses the following formulas:\\
$\cdot$ Strain (also contraction ratio) (\%)
\begin{equation}\label{eqn}
\frac{x}{L_{initial}}
\end{equation}
$\cdot$ Strain rate ($\rm {\%\cdot s^{-1}}$)
\begin{equation}\label{eqn}
\frac{\delta x}{L_{initial} \delta t}
\end{equation}
$\cdot$ Actuation stress (MPa)
\begin{equation}\label{eqn}
\frac{F}{A}
\end{equation}
$\cdot$ Force-to-weight ratio ($\rm{kN\cdot kg^{-1}}$)
\begin{equation}\label{eqn}
\frac{F}{m}
\end{equation}
$\cdot$ Specific force-to-weight ratio ($\rm {kN\cdot kg^{-1}\cdot kPa^{-1}}$)
\begin{equation}\label{eqn}
\frac{F}{m\Delta P}
\end{equation}
$\cdot$ Force-to-volume ratio ($\rm{kN\cdot m^{-3}}$)
\begin{equation}\label{eqn}
\frac{F}{V_{flat}}
\end{equation}
$\cdot$ Specific force-to-volume ratio ($\rm{kN\cdot m^{-3}\cdot kPa^{-1}}$)
\begin{equation}\label{eqn}
\frac{F}{V_{flat}\Delta P}
\end{equation}
$\cdot$ Power (kW)
\begin{equation}\label{eqn}
W=F\frac{\delta h}{\delta t}
\end{equation}
$\cdot$ Power density ($\rm{kW\cdot kg^{-1}}$)
\begin{equation}\label{eqn}
\frac{W}{m}
\end{equation}
$\cdot$ Power-to-volume ratio ($\rm{kW\cdot m^{-3}}$)
\begin{equation}\label{eqn}
\frac{W}{V_{flat}}
\end{equation}
$\cdot$ Specific work ($\rm{kJ\cdot kg^{-1}}$)
\begin{equation}\label{eqn}
\frac{m_L gh}{m}
\end{equation}
$\cdot$ Work density ($\rm{kJ\cdot m^{-3}}$)
\begin{equation}\label{eqn}
\frac{m_L gh}{V_{flat}}
\end{equation}
$\cdot$ Mechanical energy conversion efficiency (\%)
\begin{equation}\label{eqnEfficiency}
\eta=\frac{E_{out}}{E_{in}} =\frac{m_L gh}{\int_{0}^{t}q\Delta Pdt}
\end{equation}

\section{Contraction resistance measurement}
\label{appResistance}
\begin{figure}
    \centering
    \includegraphics[width=0.485\textwidth]{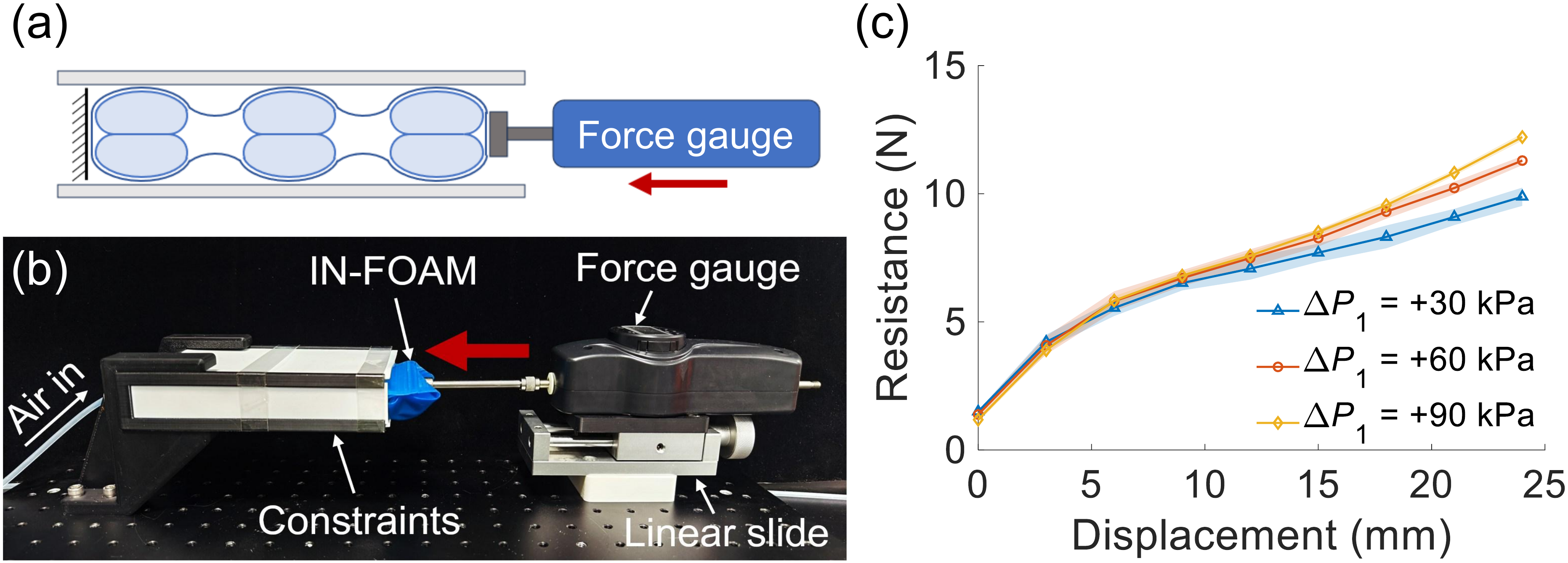}
    
    \caption{Compressive resistance measurement of the linear IN-FOAM. (a) Schematic of the compressive resistance measurement method. (b) Experiment setup for the compressive resistance measurements. (c) Relations between the compressive resistance of an inflated IN-FOAM and the displacement.}

    \label{figResistance}
\end{figure}

We estimated the resistance coefficient, $k_r$, of IN-FOAMs by measuring the compressive stiffness of the inflated skeleton (Fig. \ref{figResistance} (a-c)). During the measurement, the inflated IN-FOAM was placed in an open box made of PTFE plates (to decrease the friction brought to the IN-FOAM during contraction) and a PLA frame. A force gauge was mounted on a linear slide (Fig. \ref{figResistance}(b)), and as the slide moved, the gauge head pushed the IN-FOAM slowly. The gauge head was adjusted to make contact with the IN-FOAM. This position was defined as the origin of displacement. The force gauge head was then moved 3 mm per step, and the resistance of the IN-FOAM at each position was recorded.

{
\renewcommand{\arraystretch}{1.5}
\begin{table}
 \centering
 {\fontfamily{ptm}\selectfont
 \caption{Parameters in finite element (FE) simulations.}
  \begin{tabular}[htbp]{@{}ccc@{}}
    \hline
    Parameter & Value \\
    \hline
    Skeleton size & $23\times10\times0.012 \:\rm{cm^3}$\\
    Skin size & $28\times10.6\times0.017 \:\rm{cm^3}$\\
    Young's modulus & 400 MPa \\
    Poisson's ratio & 0.3\\
    Initial plastic strain & 0.04\\
    Contact stiffness & 1 $\rm {N\cdot mm^{-3}}$\\
    Contact damping & $10^{-4}\ \rm{ N\cdot s\cdot  mm^{-3}}$\\
    Frictional coefficient & 0.1\\
    Element type  & 3-node and 4-node shell elements\\
    Element size & 2.2 $\rm {mm^{2}}$\\
    \hline
  \end{tabular}
  \label{tabSimulation}
  }
\end{table}
}

The skeleton was inflated to +30 kPa, +60 kPa, and +90 kPa in three sets of measurements, each repeated three times. The results are shown in Fig. \ref{figResistance}(c). For displacements less than 9 mm, the resistance is similar under different pressures, and the relations between resistance and displacement are nonlinear, likely due to resistance arising from the compression of sheet materials. For displacements greater than 9 mm, the linearity of the resistance-displacement relationship increases, and the slope increases as the skeleton pressure increases. Linear fitting was performed, and the slopes of the fitted lines multiplied by $(n-1)L_{20}$ (sum of the theoretical maximum compressive displacement between the skeleton columns) were taken as the maximum resistance during the contraction, $F_{r,\:max}(\Delta P_1)$. Then contraction resistance coefficient $k_r(\Delta P_1)$ was obtained from $F_{r,\:max}(\Delta P_1)$ divided by $n(1-2/\pi)L_{10}+(n-1)L_{20}$ (the theoretical maximum contraction of the actuator).

\section{Simulation setups}
\label{appSimulation}
The finite element (FE) simulations of the linear IN-FOAM are developed using ABAQUS software. To reduce the computational cost, only one-quarter of the muscle is modeled, and the symmetric boundary conditions are applied. The model is assembled by six layers of sheets, and those areas bounded by the heat-pressing processes are constrained together.

Elastoplastic properties are assigned for the materials, and the initial plastic strain is uniformly applied to the skeleton and the skin. The elastic behavior is roughly in accord with the material’s tensile test (using INSTRON 68TM-10), when the strain is not too large. The plastic strain may be induced by the heat-pressing process of the thermoplastic materials, and this can be proved by measuring the change in a sample’s length before and after the heat-pressing process.

\begin{figure}
    \centering
    \includegraphics[width=0.485\textwidth]{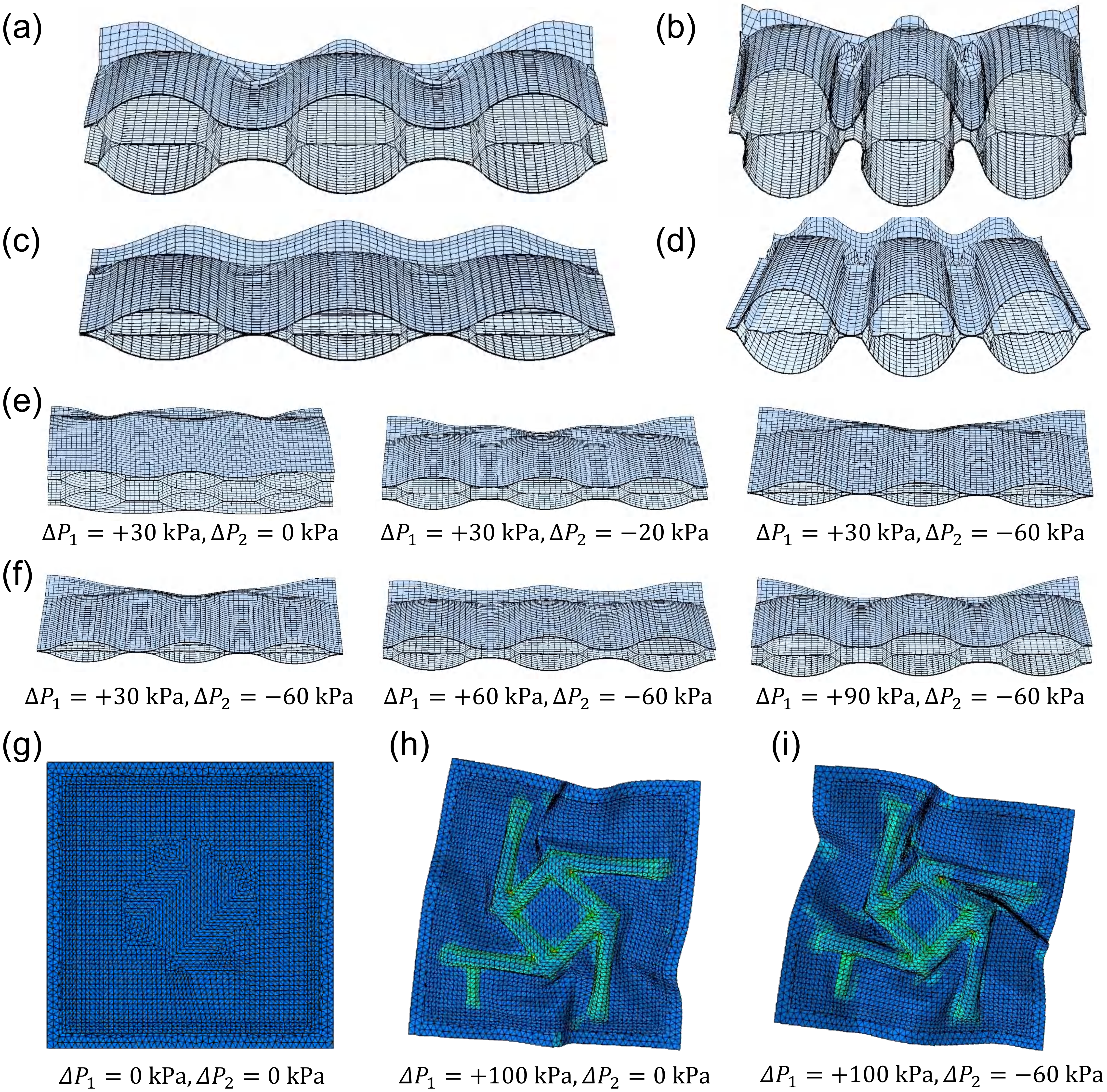}
    
    \caption{Finite element simulations of IN-FOAMs. (a-d) Cross-sectional views of the simulation results corresponding to Models A, B, C, and D, respectively. Shallow colors are used for the clarity of the cross-sections. (e) Cross-sectional views when the negative pressure increases, with the positive pressure fixed at +30 kPa. (f) Cross-sectional views when the positive pressure increases, with the negative pressure fixed at -60 kPa. (g-i) Simulation results of the IN-FOAM with a flasher-origami-like skeleton. The overall motion and wrinkles in our simulation resemble the experimental results (Fig. \ref{figMotions}(d)).}

    \label{figSimulation}
\end{figure}

\begin{figure}[t]
    \centering
    \includegraphics[width=0.485\textwidth]{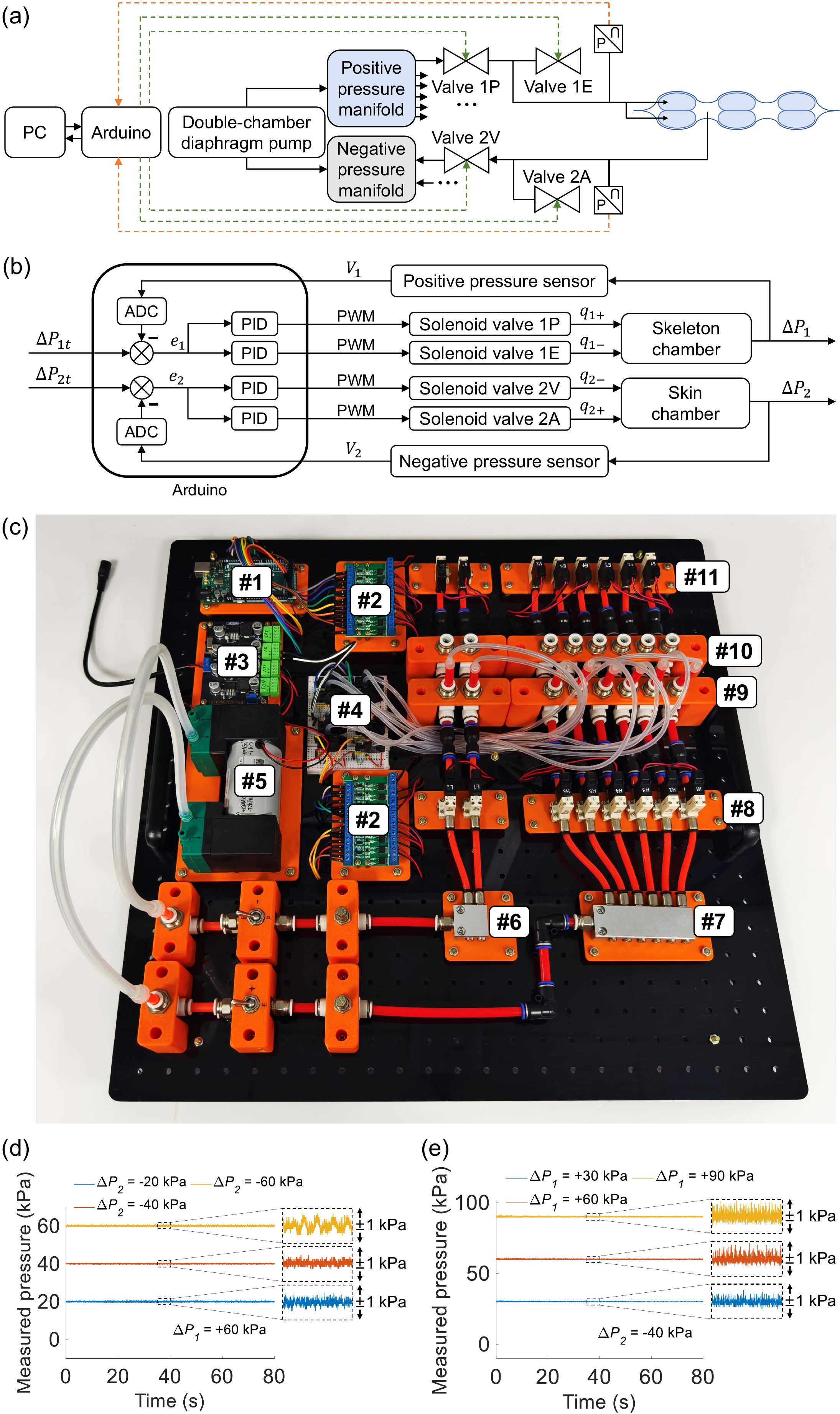}
    
    \caption{Pressure control system built in this work. (a) Schematic of the pressure control system. The Arduino board sends PWM control signals (green dashed lines) to the valves. The pressures in pneumatic circuits are measured by sensors, and the signals are transmitted to the Arduino board (orange dashed lines) as feedback. (b) Control block diagram of the system. Definitions of the symbols: $\Delta P_t$-target pressure, $e$-error between target and actual pressure, $q$-flow rate, $\Delta P$-actual pressure, $V$-output voltage of the sensors. (c) A photo of the pressure control system. Labels: \#1-Arduino microcontroller board, \#2-relays, \#3-power module, \#4-pressure sensors, \#5-diaphragm pump, \#6-negative pressure manifold, \#7-positive pressure manifold, \#8-solenoid valves for pressure-supply control, \#9-push-to-connect tube fittings to pressure sensors, \#10-push-to-connect tube fittings to external loads, \#11-solenoid valves for vent control. (d) Measured negative pressure curves during the force-contraction tests. The positive pressure was kept around +60 kPa, and the negative pressures fluctuated within ±1 kPa of the target values under closed-loop control. (e) Measured positive pressure curves during the force-contraction tests. The negative pressure was kept around -40 kPa, and the positive pressures fluctuated within ±1 kPa of the target values under closed-loop control.}

    \label{figControl}
\end{figure}

The contacts between different layers of the sheets, as well as their self-contacts, may occur at different stages of the contraction. The normal contact force is applied using the spring-damper model, which is large enough to prevent unrealistic penetrations. A fine mesh is set to capture the detailed deformations, and the total number of elements is nearly 20,000. The explicit dynamics solver is used to ease the computation for contacts, rather than the implicit quasi-static solver. As a consequence, the calculated force during contraction would have slight oscillations when buckling and local wrinkles suddenly occur. This may be aided by further considering visco-elasticity, with the cost of a more computational burden. Parameters in our simulation are provided in Table \ref{tabSimulation}.

{
\renewcommand{\arraystretch}{1.5}
\begin{table*}[!b]
 \centering
 \begin{threeparttable}
 \caption{Comparison of pneumatic sleeve haptic wearables.}
  \begin{tabular}[!b]{@{}cccccccc@{}}
    \hline
    Reference & Haptic stimuli & \makecell{Number of\\motion modes} & Low-profile & \makecell{Number of\\actuators} & \makecell{Weight of\\device} & Actuator material & Sleeve material \\
    \hline
    This work & I/K & 4 & Yes & 4 & 52 g & Heat-sealable fabric & Heat-sealable farbic \\
    \cite{du2024haptiknit} & C/I & $-$ & No & 8 & 440 g & Elastomeric resin & Knit fabric \\
    \cite{zhang2024haptic} & C/I & $-$ & Yes & 40 & 67 g & TPU film & TPU film \\
    \cite{jumet2023fluidically} & I & $-$ & Yes & 6 & 40 g & Heat-sealable fabric & Heat-sealable fabric\\
    \cite{liu2021thermocaress} & I/T & $-$ & No & 5/3 & Unknown & Unknown & Fabric \\
    \cite{zhu2020pneusleeve} & C/S/V & $-$ & No & 6 & 26 g & Woven fabric; latex tube & Knit fabric \\
    \cite{wu2019wearable} & I & $-$ & Yes & 6 & Unknown & Thermoplastic sheet & Cotton \\
    \cite{ang2019design} & K & 2 & No & 2 & Unknown & TPU & Fabric \\
    \cite{realmuto2019robotic} & K & 2 & No & 2 & 200 g & Fabric; silicone tube & Elastomer\\
    \cite{park2019lightweight} & K & 2 & Yes & 2 & Unknown & Heat-sealable fabric & Fabric \\
    \cite{zhu2017carpal} & K & 2 & Yes & 2 & Unknown & TPU film & Fabric \\
    \hline
  \end{tabular}
  \label{tabWearable}
  \begin{tablenotes}[flushleft]
    \footnotesize
    \item \textit{Note:} Abbreviations - C=compression, I=indentation, K=kinesthetic, S=skin stretch, T=temperature, V=vibration.
  \end{tablenotes}
  \end{threeparttable}
\end{table*}
}

\section{Constant pressure control}
\label{appPressureControl}
The air pressures in the actuators are regulated by a customized pressure control system (Fig. \ref{figControl} (a) and (b)). The positive and negative air pressures are supplied by a double-chamber diaphragm pump (maximum flow rate: 10 L$\cdot$min$^{-1}$ for one chamber, maximum positive pressure: +200 kPa, maximum negative pressure: -70 kPa). The pressure generated by the diaphragm pump is distributed into multiple pneumatic circuits via a manifold. Each pneumatic circuit's pressure is regulated through two solenoid valves (10-VQ110L, SMC Corporation). One solenoid valve controls the connection between an external chamber and the manifold, while the other controls the connection between the external chamber and the atmosphere. The air pressure in each pneumatic circuit is measured using a pressure sensor (MPX5100DP, NXP Semiconductors, B.V.), and the measured pressure signal is transmitted to a microcontroller board (Arduino Mega 2560 Rev3). Using the PID method, closed-loop control of the external chamber's pressure is achieved (Fig. \ref{figControl} (c)). This pressure control system can independently regulate the positive pressure (0 to +100 kPa) in six external chambers and the negative pressure (0 to -100 kPa) in two additional external chambers.

\begin{figure}
    \centering
    \includegraphics[width=0.485\textwidth]{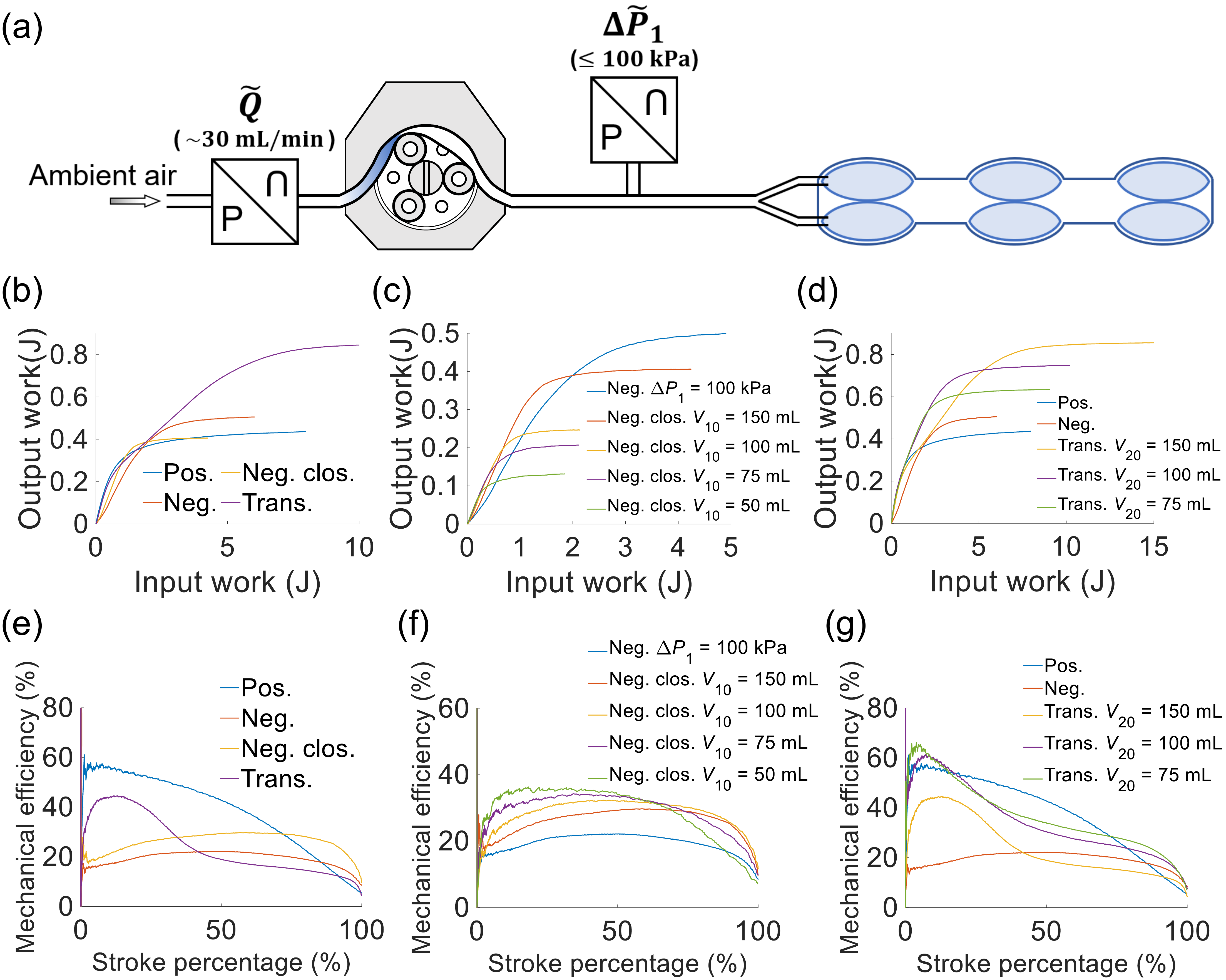}
    
    \caption{energy conversion efficiency tests of the linear IN-FOAM. Pos.: positive pressure actuation; Neg.: negative pressure actuation; Neg. clos.: negative pressure actuation with closed-skeleton; Trans.: transferring air from the skin to the skeleton. (a) Schematic of the experiment setup in energy conversion efficiency tests. (b) Relations between input and output work under different operation conditions (the initial amount of air in the closed chambers is 150 mL). (c) Relations between input and output work as the amount of air in the closed-skeleton changes. (d) Relations between input and output work when the actuator is actuated by positive pressure, negative pressure, and transferring air from the skin to the skeleton. (e) Average mechanical energy conversion efficiencies during the actuator's contraction under different operation conditions (the initial amount of air in the closed chambers is 150 mL). (f) Average mechanical energy conversion efficiencies during the actuator's contraction as the amount of air in the closed-skeleton changes. (g) Average mechanical energy conversion efficiencies during the actuator's contraction when the actuator is actuated by positive pressure, negative pressure, and transferring air from the skin to the skeleton. }

    \label{figEnergy}
\end{figure}

\section{Step response tests}
\label{appPowerDensity}
In the step response tests, an air compressor (maximum flow rate: 32 L$\cdot$min$^{-1}$, working pressure: +700 kPa) and a vacuum pump (maximum flow rate: 330 L$\cdot$min$^{-1}$, maximum negative pressure: -99 kPa) provided positive and negative pressures to the artificial muscles via pressure regulators (IR2000-A for positive pressure, and IRV20 for negative pressure, SMC Corporation). A solenoid valve (3V308NC for negative pressure, and 2SAL050 for positive pressure, AIRTAC International Group) controls the connection between the air source and the external chamber. When the power supply is initialized, the solenoid valve is activated, connecting the external chamber to the air source, making the artificial muscle contract and lift the load. A marker is affixed to the load, whose motion is recorded using a camera at 60 FPS, and is then tracked with the Tracker software. When the solenoid valve was activated, an LED was lit up, which was captured by the camera and used as an indicator of initializing time.


\section{Energy Conversion Efficiency Tests}
\label{appEnergy}
A peristaltic pump was used to pump air into the linear IN-FOAM from the atmosphere (Fig. \ref{figEnergy} (a)), while the linear IN-FOAM was fixed on a stand with a 1 kg load. The volumetric flow rate of the peristaltic pump, and the pressure difference between the pneumatic circuit and the atmosphere were measured using flow and pressure sensors (PMF2101V for flow rate, POSIFA Technologies; 40PC015G for positive pressure, and 40PC015V for negative pressure, Honeywell International Inc.). The energy conversion efficiency was calculated using Equation (\ref{eqnEfficiency}). When air was pumped from the skin to the skeleton, the pressure difference between the skeleton chamber and the skin chamber was used for $\Delta P$.

\begin{figure}
    \centering
    \includegraphics[width=0.485\textwidth]{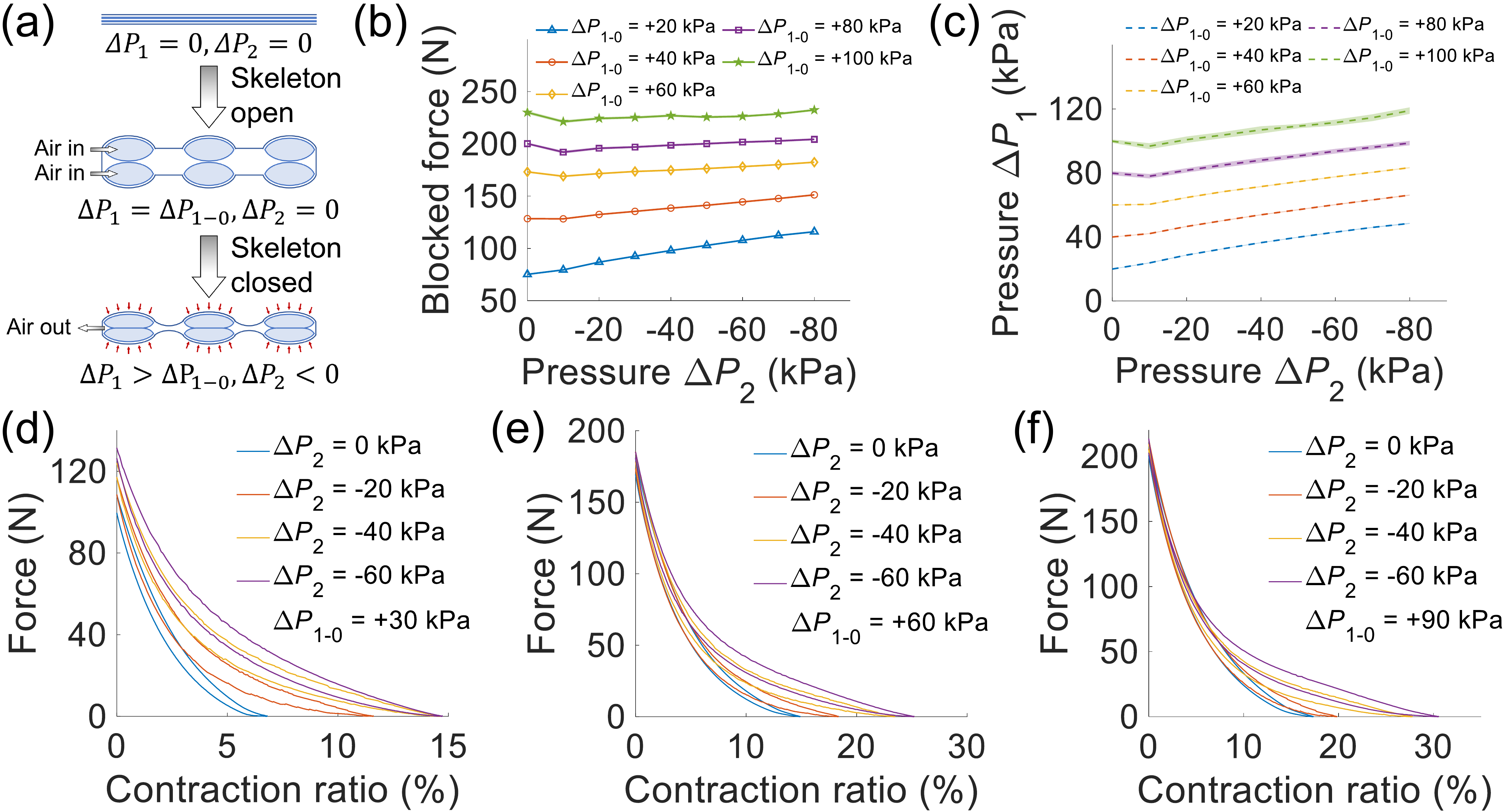}
    
    \caption{Static property of the linear IN-FOAM with closed skeleton. (a) Illustration of the skin pressing to the closed skeleton under negative pressure $\Delta P_2$. $\Delta P_{1-0}$: the initial value of $\Delta P_1$ before closing the skeleton and applying vacuum to the skin. (b) Blocked force as a function of $\Delta P_2$ under different $\Delta P_1$. (c) Variation of the positive pressure in the closed skeleton as $\Delta P_2$ increases. (d-f) Force-contraction curves when $\Delta P_{1-0}$ equals +30 kPa, +60 kPa, and +90 kPa, respectively.}

    \label{figStatic2}
\end{figure}

\begin{figure}[ht]
    \centering
    \includegraphics[width=0.485\textwidth]{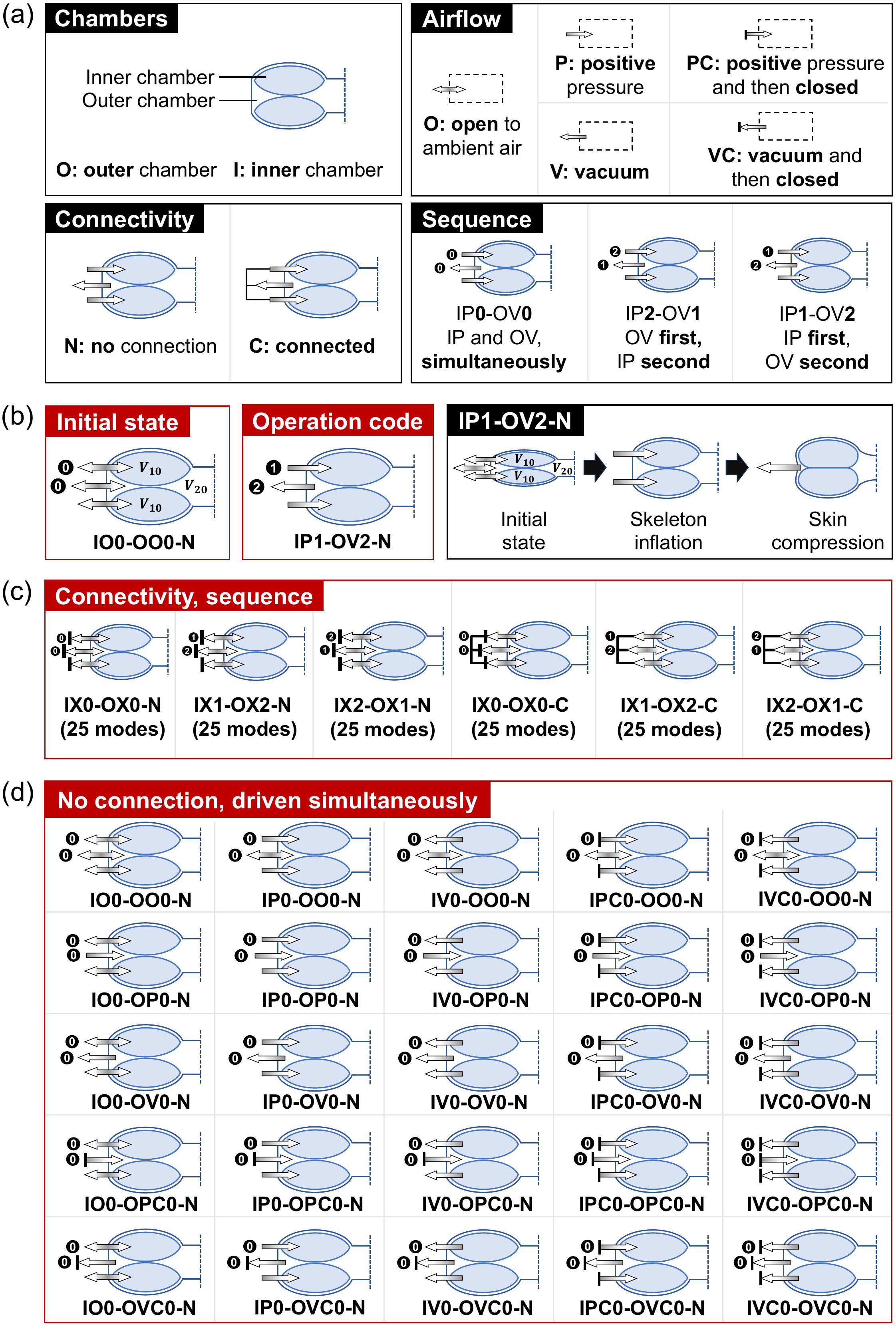}
    
    \caption{Multiple operations of IN-FOAMs. By this classification method, there are totally 150 operation modes: $5\;(\rm{skin\;airflow})\times5\;(\rm{skeleton\;airflow})\times2\;(\rm{connectivity})\times3\;(\rm{pressurizing\;sequences})$. (a) Code definitions of different operation conditions including airflow, connectivity of the skin and the skeleton, and pressurizing sequences. (b) Initial state of the actuator and one example of the operation code. The initial state is that the inner and outer chambers are open to ambient air. The code IP1-OV2-N means that the inner chamber is inflated first, the outer chamber is secondly driven by vacuum, and the inner and outer chambers are not connected. (c) Operation conditions taking connectivity and pressurizing sequences into consideration. (d) Operation conditions when the skin and the skeleton chambers are independent and driven simultaneously.}

    \label{figOperations}
\end{figure}

We calculated the energy conversion efficiency as a function of the actuator's stroke percentage. During the working process of the actuator, the energy conversion efficiency is not constant. From the output-input work curve (Fig. \ref{figEnergy} (b-d)), it can be observed that the energy conversion efficiency of the linear IN-FOAM (represented by the slope of the secant line starting from the origin) initially increases and then decreases as the actuator contracts. During the positive pressure actuation process, the energy conversion efficiency gradually decreases from its maximum value of 57\:\% and eventually becomes 5.5\:\% (Fig. \ref{figEnergy} (e)). In the negative pressure actuation process, the energy conversion efficiency remains above 15\:\% during the first 90\:\% stroke and finally decreases to 8.4\:\% (Fig. \ref{figEnergy} (e)).

By altering the operation conditions of IN-FOAMs, the energy conversion efficiency curve during the operation can be tuned. Pre-inflating the skeleton with a certain amount of air and then closing it makes the overall energy conversion efficiency curve rise during the negative pressure actuation process. As the pre-inflation volume increases, the energy conversion efficiency decreases for the first 70\:\% stroke, and increases for the final 30\:\% stroke (Fig. \ref{figEnergy} (f)), with an increase of the stroke.

We also explored transferring air from the skin chamber to the skeleton chamber, which increased the negative pressure in the skin and the positive pressure in the skeleton at the same time. Under this condition, the energy conversion efficiency curve rises sharply at the beginning of contraction, followed by a rapid decline, and then stabilizes with a gentle downward slope. As the pre-inflation volume of the skin increases (Fig. \ref{figEnergy} (g)), the energy conversion efficiency drops, accompanied by an increase in the stroke.

Under positive pressure actuation, the peak energy conversion efficiency is approximately 57\:\%, with an average energy conversion efficiency of 5.5\:\% over the entire stroke. Under negative pressure actuation, the peak energy conversion efficiency is around 22\:\%, with an average energy conversion efficiency of 8.4\:\% over the entire stroke. Pre-inflating the skeleton and then closing the skeleton makes the peak energy conversion efficiency under negative pressure actuation increase to 36\:\% and the average energy conversion efficiency increase to 11.5\:\%; however, this reduces the stroke. Pre-inflating the skin chamber and driving the linear IN-FOAM by transferring air from the skin to the skeleton makes the maximum energy conversion efficiency increase to 64\:\%, with an increase in stroke. In practical use, an appropriate operation condition can be selected to enhance energy conversion efficiency while meeting the stroke requirements.

\section{Multiple Optional Operation Modes}
\label{appMultipleOperation}
A single-mode IN-FOAM can be regarded as a pneumatic system with two input chambers. By varying the operation conditions of the two chambers, IN-FOAMs can generate diverse contraction processes and exhibit a wide range of characteristics. These operation conditions include the airflow direction, the pressurizing sequences, and whether the two chambers are connected. A detailed classification of these operation modes is shown in Fig. \ref{figOperations} (a), along with their representative operation codes. Assuming the initial state is that both the skin and the skeleton are connected to the atmosphere, a simple combinatorial calculation indicates that IN-FOAMs have a total of 150 operation modes under this classification method (Fig. \ref{figOperations} (c) and (d)). Each mode can be represented by a unique combination code (Fig. \ref{figOperations} (b)).

However, not all of these operational modes enable IN-FOAMs to function as artificial muscles. For instance, extracting air from the skeleton while inflating the skin will prevent normal contraction. In this case, as the skin inflates, the entire IN-FOAM becomes an airbag, and can act as an elastic damper in a pneumatic system. Another example is simultaneously extracting air from both the skin and skeleton, which makes the IN-FOAM return to a flat state without functionality. Our current research focuses primarily on operation modes including IP1-OO2-N, IPC1-OO2-N, IP1-OV2-N, IPC1-OV2-N, IP0-OV0-N, and IP0-OV0-C. The functionality and characteristics of other modes require further exploration.


\bibliographystyle{ieeetr}
\bibliography{references}


\newpage

\begin{IEEEbiography}[{\includegraphics[width=1in,height=1.25in,clip,keepaspectratio]{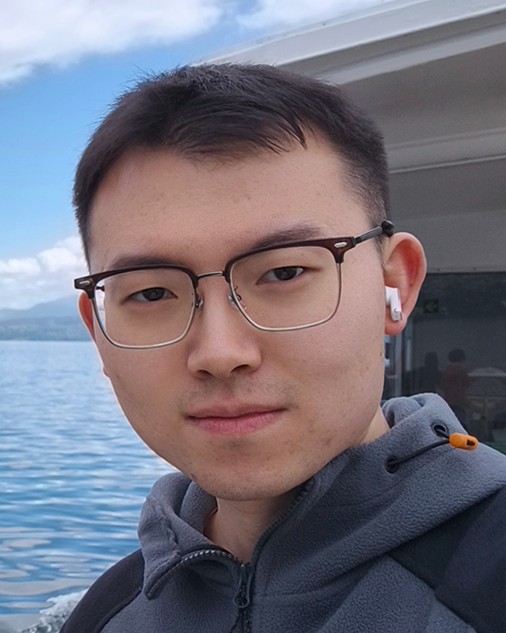}}]{Siyuan Feng}
received the B.E. degree in Mechanical Engineering (with honors) and a minor in Automation from Tsinghua University (Beijing, China) in 2024. He is now working toward the Ph.D. degree in Mechanical Engineering in the Department of Mechanical Engineering at Tsinghua University (Beijing, China). His research interests include soft robotics, computational design, and physics-based simulation.
\end{IEEEbiography}

\vspace{-300pt}

\begin{IEEEbiography}[{\includegraphics[width=1in,height=1.25in,clip,keepaspectratio]{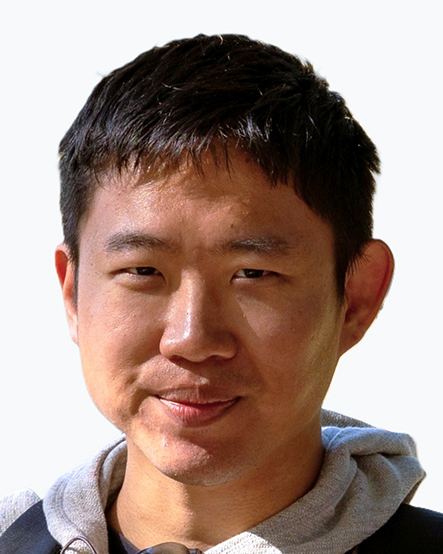}}]{Ruoyu Feng}
received the B.S. degree in Engineering Mechanics from Shandong University (Jinan, China) in 2018, and the Ph.D. degree in Mechanics from Tsinghua University (Beijing, China) in 2023. He is currently a postdoctoral researcher in the Department of Mechanical Engineering at Tsinghua University (Beijing, China). His research interests include soft robotics, bioinspired robots, and multi-body dynamics.
\end{IEEEbiography}

\vspace{-300pt}

\begin{IEEEbiography}[{\includegraphics[width=1in,height=1.25in,clip,keepaspectratio]{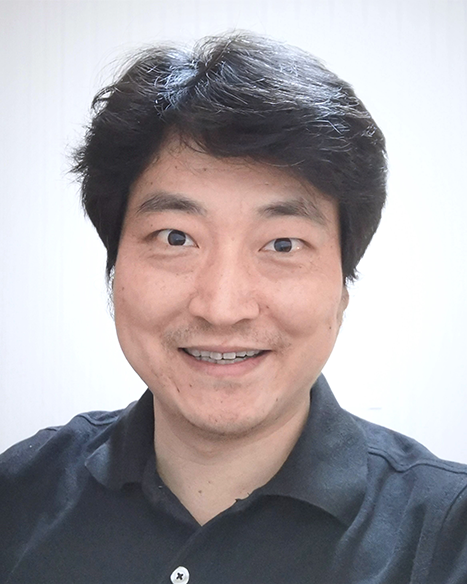}}]{Shuguang Li}
is currently an Associate Professor in the Department of Mechanical Engineering at Tsinghua University (Beijing, China). He earned his Ph.D. in Mechanical and Aerospace Engineering from Northwestern Polytechnical University (Xi’an, China) in 2013. Before joining Tsinghua, he held positions as a Postdoctoral Fellow and Research Associate at Cornell University, the Massachusetts Institute of Technology (MIT), and Harvard University. His current research interests include collective robotics and soft robotics.
\end{IEEEbiography}
 
%

\newpage

 




\vfill

\end{document}